\documentclass[manuscript,screen]{acmart}

\AtBeginDocument{%
  }

\setcopyright{acmlicensed}
\copyrightyear{2026}
\acmYear{2026}
\acmDOI{XXXXXXX.XXXXXXX}

\usepackage{amsmath,amsfonts}
\usepackage{algorithmic}
\usepackage{algorithm}
\usepackage{array}
\usepackage{textcomp}
\usepackage{stfloats}
\usepackage{url}
\usepackage{verbatim}
\usepackage{graphicx}
\usepackage{cite}
\usepackage{orcidlink}
\usepackage{tabularx}
\usepackage[table]{xcolor}
\usepackage{booktabs}
\usepackage{caption}
\usepackage{multirow}
\usepackage{makecell}
\usepackage[caption=false,font=footnotesize]{subfig}
\usepackage{pifont}
\newcolumntype{H}{>{\setbox0=\hbox\bgroup}c<{\egroup}@{}}

\usepackage{tcolorbox}
\tcbuselibrary{skins}

\newtcolorbox{systempromptbox}[1]{
    enhanced,
    colback=white,
    colframe=blue!50!black,
    fonttitle=\bfseries,
    title=#1,
}

\newtcolorbox{userpromptbox}[1]{
    enhanced,
    colback=white,
    colframe=green!50!black,
    fonttitle=\bfseries,
    title=#1,
}

\definecolor{romanticism_color}{HTML}{756bb1}
\definecolor{expressionism_color}{HTML}{ffbb78}
\definecolor{minimalism_color}{HTML}{c7c7c7}
\definecolor{abstractart_color}{HTML}{266a9b}
\definecolor{abst_expressionism_color}{HTML}{ff7f0e}
\definecolor{artnouveau_color}{HTML}{8c564b}
\definecolor{earlyreinass_color}{HTML}{aec7e8}
\definecolor{symbolism_color}{HTML}{e6550d}

\newcolumntype{M}[1]{>{\centering\arraybackslash}m{#1}}

\usepackage{soul}

\newcommand{\ArtGraph}{\mathcal{A}\textit{rt}\mathcal{G}\textit{raph}}

\newcommand{\ArtSound}{\mathcal{A\textit{rt}S\textit{ound}}}

\newcommand{\ArtToMus}{\mathcal{A}\textit{rt2}\mathcal{M}\textit{us}}

\newcommand{\ICscore}{\mathcal{ICS}\textit{core}}
\newcommand{\ACscore}{\mathcal{ACS}\textit{core}}

\begin{document}

\title{$\ArtToMus$: Artwork-to-Music Generation via Visual Conditioning and Large-Scale Cross-Modal Alignment}

\author{Ivan Rinaldi}
\orcid{0009-0003-9597-064X}
\affiliation{%
  \institution{University of Bari Aldo Moro}
  \city{Bari}
  \country{Italy}
}
\email{i.rinaldi4@phd.uniba.it}

\author{Matteo Mendula}
\orcid{0000-0002-0126-9808}
\affiliation{%
  \institution{Catalonia's Telecommunications Technology Centre}
  \city{Barcelona}
  \country{Spain}
}
\email{matteo.mendula@cttc.cat}

\author{Nicola Fanelli}
\orcid{0009-0007-6602-7504}
\affiliation{%
  \institution{University of Bari Aldo Moro}
  \city{Bari}
  \country{Italy}
}
\email{nicola.fanelli@uniba.it}

\author{Florence Levé}
\orcid{0000-0002-6571-5096}
\affiliation{%
  \institution{University of Picardie Jules Verne}
  \city{Amiens}
  \country{France}
}
\email{florence.leve@u-picardie.fr}

\author{Matteo Testi}
\orcid{0009-0004-3266-0457}
\affiliation{%
 \institution{Artificial Intelligence Venture Builder (AIVB)}
 \city{London}
 \country{England}
}
\email{matteo.testi@aivb.ai}

\author{Giovanna Castellano}
\orcid{0000-0002-6489-8628}
\affiliation{%
  \institution{University of Bari Aldo Moro}
  \city{Bari}
  \country{Italy}
}
\email{giovanna.castellano@uniba.it}

\author{Gennaro Vessio}
\orcid{0000-0002-0883-2691}
\affiliation{%
  \institution{University of Bari Aldo Moro}
  \city{Bari}
  \country{Italy}
}
\email{gennaro.vessio@uniba.it}

\renewcommand{\shortauthors}{Rinaldi et al.}

\begin{abstract}
Music generation has advanced markedly through multimodal deep learning, enabling models to synthesize audio from text and, more recently, from images. However, existing image-conditioned systems suffer from two fundamental limitations: (\textit{i}) they are typically trained on natural photographs, limiting their ability to capture the richer semantic, stylistic, and cultural content of artworks; and (\textit{ii}) most rely on an image-to-text conversion stage, using language as a semantic shortcut that simplifies conditioning but prevents direct visual-to-audio learning. Motivated by these gaps, we introduce $\ArtSound$, a large-scale multimodal dataset of 105,884 artwork–music pairs enriched with dual-modality captions, obtained by extending $\ArtGraph$ and the Free Music Archive. We further propose $\ArtToMus$, the first framework explicitly designed for \textit{direct} artwork-to-music generation, which maps digitized artworks to music without image-to-text translation or language-based semantic supervision. The framework projects visual embeddings into the conditioning space of a latent diffusion model, enabling music synthesis guided solely by visual information. Experimental results show that $\ArtToMus$ generates musically coherent and stylistically consistent outputs that reflect salient visual cues of the source artworks. While absolute alignment scores remain lower than those of text-conditioned systems—as expected given the substantially increased difficulty of removing linguistic supervision—$\ArtToMus$ achieves competitive perceptual quality and meaningful cross-modal correspondence. This work establishes direct visual-to-music generation as a distinct and challenging research direction, and provides resources that support applications in multimedia art, cultural heritage, and AI-assisted creative practice.
\end{abstract}




\maketitle

\section{Introduction}

Music generation has long been one of the most expressive forms of computational creativity. Music combines structure, emotion, and cultural background, making its automatic generation intrinsically challenging. Translating visual artworks into music is even more demanding, not because the two modalities are unrelated, but because their correspondences are indirect and highly structured. Visual art encodes meaning through composition, color, texture, style, and historical context, evoking emotions and narratives that are central to musical expression~\citep{zhao2020emotion}. These shared perceptual and affective dimensions suggest meaningful cross-modal correspondences between images and sound. However, such relationships are rarely one-to-one or directly observable; they emerge from higher-level abstractions shaped by human perception and artistic conventions, making them difficult to model without sufficiently expressive representations.

Recent advances in multimodal deep learning have substantially improved generative models, enabling the synthesis of symbolic music~\citep{shih2022theme, hu2023beauty, ma2024symbolic} and natural audio from text, images, or multimodal inputs~\citep{agostinelli2023musiclm, copet2023simple, liu2023m}. Text-based conditioning provides strong semantic grounding, and Large Language Models (LLMs) further reinforce this paradigm by converting heterogeneous inputs into rich textual descriptions~\citep{liu2026mumu, liu2024audioldm}. However, two limitations remain.

First, most image-conditioned music generation systems are trained primarily on natural photographs, whose semantics are comparatively literal and object-centric. Such data only partially reflect the symbolic, stylistic, and culturally layered semantics of artworks, limiting the applicability of existing approaches to the artistic domain.

Second, many pipelines rely on an image-to-text conversion stage. While effective, this strategy introduces language as an intermediate semantic bottleneck. Linguistic descriptions inevitably abstract away ambiguity, texture, and stylistic details—elements that often resist concise verbalization but are central to artistic expression. Consequently, image-to-music systems that depend on textual intermediaries may produce musically coherent outputs while failing to preserve the richer expressive structure of the source artwork. In this sense, language acts as a semantic shortcut: it simplifies multimodal conditioning by introducing explicit, human-defined concepts, at the cost of discarding potentially relevant non-verbalizable visual cues relevant to musical generation.

Existing attempts at image-to-music generation illustrate these issues. Early methods relied on handcrafted image features coupled with RNN-based decoders~\citep{tan2020automated}, resulting in limited expressive variability. In~\citep{hisariya2024bridging}, artwork images are converted into emotional descriptors and captions that guide music generation. More recent approaches~\citep{song2025musflow, li2025mozart} leverage Multimodal Large Language Models (MLLMs) to describe images and subsequently employ text-to-music generators. While effective, such pipelines delegate the core multimodal alignment to the language model and cannot learn direct relationships between visual semantics and musical structure. As a consequence, the dependency between visual structure and generated music is mediated by linguistic abstractions rather than inferred directly from visual data. Furthermore, progress in this direction is constrained by the scarcity of large-scale datasets explicitly pairing artworks with music. Whereas extensive resources exist for text-to-image~\citep{wang2023diffusiondb, srinivasan2021wit} and text-to-audio~\citep{bai2025audiosetcaps, agostinelli2023musiclm}, only small artwork--music collections have been explored~\citep{rinaldi2024art2mus}, limiting systematic investigation of direct cross-modal learning.

Motivated by these gaps, this work addresses two necessary conditions for advancing artwork-to-music generation: (\emph{i}) enabling systematic research through the construction of a large-scale multimodal dataset; and (\emph{ii}) introducing a framework capable of generating music \emph{directly} from digitized artworks, without textual intermediates. This formulation defines a strictly harder multimodal problem. Unlike text-conditioned systems, which access explicit semantic supervision, an image-only model must infer affective, stylistic, and compositional cues directly from visual embeddings and translate them into meaningful acoustic patterns.

To support this effort, we introduce $\ArtSound$, a large-scale dataset of 105{,}884 artwork--music pairs enriched with image and audio captions validated through two new alignment metrics. Building on this resource, we propose $\ArtToMus$, the first system explicitly designed for direct artwork-to-music generation. The framework injects visual embeddings extracted from artworks into the conditioning space of a latent diffusion model through a dedicated image-to-audio interface. By aligning CLIP~\citep{radford2021learning} or ImageBind~\citep{girdhar2023imagebind} embeddings with GPT-2's~\citep{radford2019language} representation space, $\ArtToMus$ enables music synthesis guided solely by visual information. This design intentionally removes language-based semantic supervision, forcing the model to learn cross-modal correspondences directly from visual representations rather than relying on linguistic abstractions.

In our preliminary work~\citep{rinaldi2024art2mus}, we conditioned a text-to-audio model directly on images. That study exposed two limitations addressed here: (\emph{i}) simple linear projections were insufficient to capture meaningful artwork features; and (\emph{ii}) the dataset of 10{,}000 pairs lacked sufficient stylistic and musical diversity.

The remainder of the paper is organized as follows. Section~\ref{sec:related_work} reviews related literature. Section~\ref{sec:materials} describes the dataset and captioning pipelines. Section~\ref{sec:methodology} presents the $\ArtToMus$ architecture. Section~\ref{sec:experiments} reports experimental settings and results. Section~\ref{sec:conclusion} concludes the work.

\section{Related Work}
\label{sec:related_work}

Recent advances in generative AI have increasingly focused on modeling and integrating heterogeneous data modalities, including text, images, and audio~\citep{borsos2023audiolm, zhu2023conditional, ge2023expressive, huang2023make}. A rapidly growing research direction is \emph{conditioned music generation}, in which models synthesize music guided by external inputs. This paradigm includes text-to-music approaches~\citep{sharma2024m2m, liu2023m} as well as emerging image-to-music methods~\citep{liu2026mumu, chowdhury2024melfusion, wang2025visiontomusicgenerationsurvey}. The goal is to generate musically coherent and semantically meaningful audio that reflects the context conveyed by the conditioning modality. While text-based conditioning generally yields strong alignment due to its explicit semantic structure, image conditioning remains more challenging because visual information is abstract, multi-layered, and often underspecified.

Several models have explored image-conditioned or cross-modal music generation. IM2WAV~\citep{sheffer2023hear} generates audio aligned with one or more images by extracting CLIP~\citep{radford2021learning} embeddings and feeding them to an autoregressive token-based audio model. M2M-Gen~\citep{sharma2024m2m} generates background music for Japanese manga by using an LLM to produce textual descriptions informed by scene boundaries, dialogue, and emotions inferred from characters’ expressions. In~\citep{liaoimage}, an image is first converted into lyrics via an LLM, and a rule-based composition system then produces a musical score exported as MusicXML. MelFusion~\citep{chowdhury2024melfusion} integrates visual semantics into a diffusion-based text-to-music framework by merging DDIM-inverted image latents with evolving audio latents within decoder layers. Mozart's Touch~\citep{li2025mozart} supports images, videos, and text by first captioning visual content, transforming captions into musical descriptions, and finally generating music using a text-to-music model.

Notably, most of these systems are trained and evaluated on natural photographs or visually literal content, whose semantics are primarily object-centric and scene-based. Such data only partially reflect the symbolic abstraction, stylistic variability, and historically layered semantics characteristic of artworks. As a result, existing image-conditioned music generation approaches may not generalize effectively to the artistic domain, where meaning is often encoded through compositional structure, style, and cultural context rather than explicit objects.

Moreover, despite recent progress, most image-conditioned music generation pipelines rely on textual descriptions as an intermediate modality. Images are typically converted into captions, which text-to-audio models then process. Although this strategy enables the reuse of mature architectures, it introduces an inherent information bottleneck. Once visual content is compressed into text, attributes not captured in the caption—such as fine-grained stylistic patterns, compositional nuances, or atmospheric qualities—cannot be recovered during audio generation. From an information-theoretic perspective, the intermediate textual representation can only preserve or discard visual information, but cannot introduce visual details that were not encoded in the caption. Conditioning directly on visual features may therefore preserve stylistic and compositional cues that are otherwise filtered through linguistic abstraction.

Our preliminary study~\citep{rinaldi2024art2mus} demonstrated that direct artwork-to-music generation is feasible by conditioning a text-to-audio model on visual embeddings. However, that approach was limited by simple projection mechanisms and a relatively small dataset, motivating the development of more expressive visual–audio alignment strategies and larger-scale resources.

In addition to generative pipelines, captioning has emerged as a central mechanism for cross-modal alignment. Captioning systems aim to convert unstructured inputs—such as images or audio—into textual descriptions that capture semantic content, emotion, or stylistic attributes. Early systems, including the Audio Captioning Transformer~\citep{mei2021audio} and Midge~\citep{mitchell2012midge}, relied on conventional architectures and often produced generic outputs. More recent advances have been driven by MLLMs, which generate rich and context-aware captions for images (e.g., LLaVA~\citep{liu2023visual}, Qwen2.5-VL~\citep{bai2025qwen2}) and audio (e.g., Pengi~\citep{deshmukh2023pengi}, Qwen2-Audio~\citep{chu2024qwen2}). Emerging omni-modal architectures such as Qwen2.5-Omni~\citep{xu2025qwen2} integrate text, images, and audio within unified frameworks, enabling more sophisticated cross-modal reasoning.

These advances have been supported by the release of large-scale captioning datasets~\citep{kim2019audiocaps, chen2015microsoft, plummer2015flickr30k, gemmeke2017audio}, which provide supervision for image and audio understanding. Domain-specific captioning datasets have also been introduced. In the visual art domain, ArtCap~\citep{lu2024artcap} provides crowd-sourced descriptions for Impressionist and Expressionist paintings; the Iconclass Caption Dataset~\citep{cetinic2021iconographic} offers iconographic annotations for over 86k artworks; and ArtEmis~\citep{achlioptas2021artemis} contains emotional reactions and free-text explanations for more than 80k artworks. In the music domain, datasets such as MusCaps~\citep{manco2021muscaps}, MusicBench~\citep{melechovsky2024mustango}, and MusicEval~\citep{liu2025musiceval} provide linguistic descriptions of musical attributes, mood, genre, and structure, facilitating natural-language understanding and evaluation of music.

Despite these resources, very few datasets explicitly pair artworks with music, and those that do remain limited in scale~\citep{rinaldi2024art2mus}. As a consequence, the field remains largely oriented toward text-mediated alignment, and the systematic investigation of direct visual-to-music generation remains underexplored.

Our work addresses these gaps by jointly introducing a large-scale artwork–music dataset and a direct visual-conditioning framework that learns cross-modal correspondences without linguistic supervision. With $\ArtToMus$, we condition a latent diffusion model directly on projected visual embeddings extracted from digitized artworks, enabling expressive and efficient artwork-to-music generation grounded in visual representations rather than textual abstractions.

\section{Materials} 
\label{sec:materials}

This section introduces the construction of the proposed $\ArtSound$ dataset, a unified multimodal resource that pairs digitized artworks with music tracks, enriched with image and audio captions, and aligned via embedding-based pairing strategies. We first describe the data sources and captioning pipelines, then present the metrics used to validate caption quality, and finally discuss artwork–music pairing.

\subsection{Data Sources} 
\label{sec:data_sources}

To build a large-scale dataset for artwork-to-music generation, we began by drawing from two open-access repositories. For artworks, we used $\ArtGraph$~\citep{castellano2022leveraging}, a knowledge graph derived from WikiArt and DBpedia, containing 116{,}475 digitized artworks across 18 genres and 32 styles, with rich contextual metadata, such as artist, style, and genre. For music, we adopted the Large version of the Free Music Archive (FMA) dataset~\citep{defferrard2016fma}, which includes 106{,}574 Creative Commons-licensed 30-second tracks from 16{,}341 artists spanning 16 top-level and 161 sub-genres. After removing corrupted audio files, 105{,}884 tracks were retained.


\subsection{Caption Generation} 
\label{sec:dataset_creation}

\begin{figure*}[t]
  \centering
  \includegraphics[width=.85\linewidth]{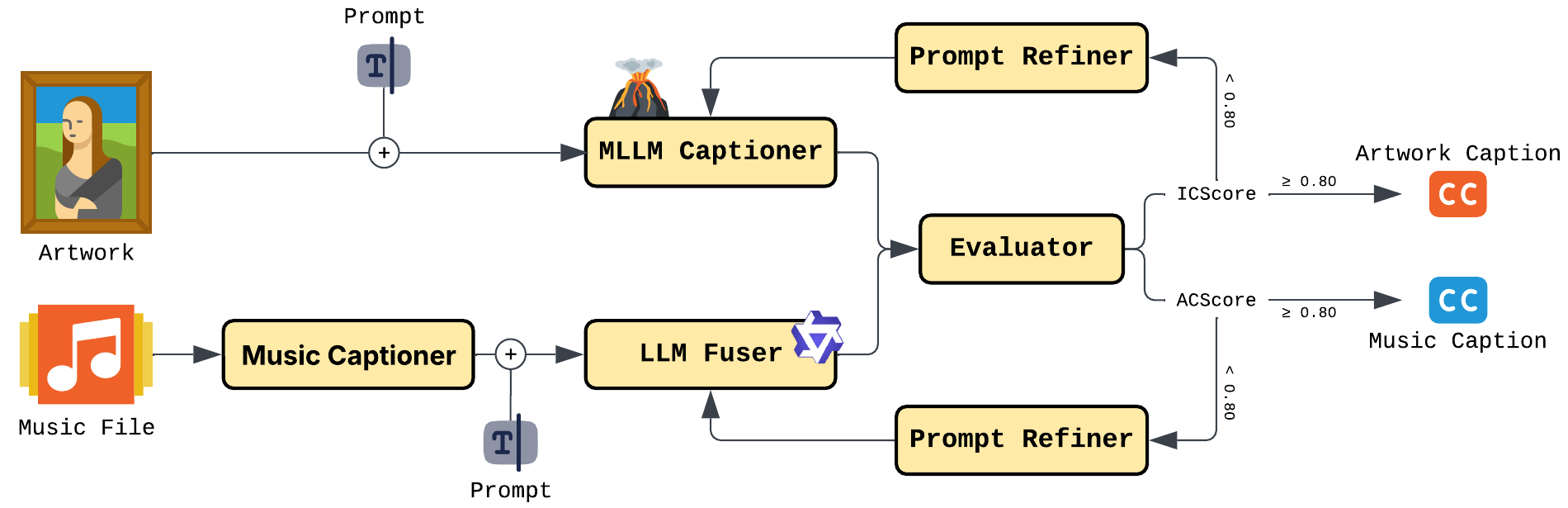}
  \caption{Unified captioning pipeline for images and audio. Images are captioned using an MLLM guided by a structured prompt; audio is captioned through a segment-level music captioner and then fused by an LLM. All captions are evaluated using the proposed $\ICscore$ and $\ACscore$ metrics; captions below threshold are regenerated with refined prompts.} 
  \Description{The figure illustrates the end-to-end captioning workflow for both images and audio, highlighting their parallel processing and convergence. It emphasizes structured prompting, segment-level fusion, and automatic quality control via iterative regeneration driven by evaluation metrics.}
  \label{fig:artw_music_caption_pipelines}
\end{figure*}

Image and audio captions were produced through a unified yet modality-specific pipeline (Figure~\ref{fig:artw_music_caption_pipelines}). Image captions were generated using LLaVA~\citep{liu2023visual}, prompted to produce rich descriptions covering artworks' content, mood, stylistic cues, and possible artistic influences. Audio captions were generated using LP-MusicCaps~\citep{doh2023lp}, which produces short 10-second segment captions; these were subsequently merged into a single coherent description using Qwen3~\citep{yang2025qwen3}.

To ensure caption reliability, both modalities were evaluated using custom multimetric scores specifically designed to reflect different aspects of alignment quality. For image captions, we defined the $\ICscore$ as a weighted combination of CLIP-Score~\citep{hessel2021clipscore} and PAC-Score~\citep{sarto2023positive}. CLIP-Score measures global semantic agreement between an image and its caption using CLIP embeddings, whereas PAC-Score evaluates whether positive semantic attributes expressed in the caption are present in the image. Combining these metrics provides a balanced assessment of both high-level semantics and fine-grained attribute consistency. Formally:
\begin{equation}
\begin{aligned}
\ICscore(x,y) &= \gamma \cdot \text{CLIP-Score}(x,y) + (1-\gamma)\cdot \text{PAC-Score}(x,y),
\end{aligned}
\label{eq:image_capt_score}
\end{equation}
where $x$ is the artwork, $y$ the generated caption, and $\gamma=0.35$ is an empirically selected weight.

For audio captions, we defined the $\ACscore$ to evaluate how well the fused caption summarizes the segment-level descriptions produced by LP-MusicCaps. ROUGE-1~\citep{lin2004rouge} captures lexical content preservation, while BERT-Score~\citep{zhang2020bertscore} measures semantic similarity in an embedding space, enabling robustness to paraphrasing. Their weighted combination yields:
\begin{equation}
\begin{aligned}
\ACscore(x,y) &= \alpha \cdot \text{ROUGE-1}(x,y) + (1-\alpha)\cdot \text{BERT-Score}(x,y),
\end{aligned}
\label{eq:aud_capt_score}
\end{equation}
where $x$ denotes the set of segment-level captions, $y$ the fused caption, and $\alpha=0.30$ controls the relative contribution of lexical vs.\ semantic overlap.

We initially considered also integrating ROUGE-L in the $\ACscore$, but it was later discarded because its emphasis on the longest common subsequence penalized valid paraphrases, lowering the score even when the fused caption was semantically consistent with the segment-level descriptions.

Captions with $\ICscore$ or $\ACscore$ below the threshold of 0.80 were regenerated using refined prompts. The refinement consisted of issuing a new prompt that explicitly informed the LLM/MLLM that the previously generated caption did not meet the required quality threshold and contained specific issues. This feedback-based prompt forced the model to regenerate the caption, explicitly ensuring compliance with the imposed requirements. The refined prompt was only applied after a first failed attempt, as using it a priori would have been logically inconsistent, given that no caption had yet been generated or evaluated at that stage. Regeneration was allowed for up to 3 iterations, as subsequent attempts showed diminishing returns. Table~\ref{tab:artsound_captions_stats} summarizes the high-level statistics related to the captions.

Even after the allowed regeneration attempts, a caption that did not satisfy the threshold was still retained in the dataset. Additionally, the corresponding $\ICscore$ and $\ACscore$ values are included in the $\ArtSound$ dataset to provide further insight into caption quality and enable downstream filtering or analysis.

\begin{table}[b]
    \centering
    \caption{Statistics for image and audio captions in $\ArtSound$, including caption length (expressed as number of characters), and quality metrics. For image captions: CLIP-Score, PAC-Score, and the composite $\ICscore$. For audio captions: BERT-Score, ROUGE-1, and the composite $\ACscore$. Values above or below the acceptance threshold (0.80) are reported.}
    \resizebox{1.\linewidth}{!}{
    \begin{tabular}{l|cccc|cccc}
        \toprule
        \multirow{2}{*}{\textbf{Statistic}} & \multicolumn{4}{c|}{\textbf{Image Captions}} & \multicolumn{4}{c}{\textbf{Audio Captions}} \\
        & \textbf{Length} & \textbf{CLIP-Score} & \textbf{PAC-Score} & \textbf{$\ICscore$} & \textbf{Length} & \textbf{BERT-Score} & \textbf{ROUGE-1} & \textbf{$\ACscore$} \\
        \midrule
        Minimum & 234.00 & 0.3699 & 0.5165 & 0.4918 & 82.00 & 0.8474 & 0.1000 & 0.6234 \\
        Average & 950.18 & 0.7821 & 0.8431 & 0.8217 & 411.76 & 0.9321 & 0.6894 & 0.8593 \\
        Maximum & 2302.00 & 1.0000 & 1.0000 & 1.0000 & 1084.00 & 0.9940 & 0.9771 & 0.9889 \\
        \cmidrule(lr){1-9}
        $>$ Threshold & -- & -- & -- & 78{,}845 & -- & -- & -- & 95{,}837 \\
        $<$ Threshold & -- & -- & -- & 37{,}630 & -- & -- & -- & 10{,}047 \\
        \bottomrule
    \end{tabular}
    }
    \label{tab:artsound_captions_stats}
\end{table}

\begin{table*}[t]
\centering
\caption{Examples of $\ArtSound$ image captions generated using LLaVA1.6-Mistral-7B.}
\small
\begin{tabularx}{\linewidth}{@{} >{\centering\arraybackslash}m{0.25\textwidth} | >{\arraybackslash}m{0.7\textwidth} @{}}
\toprule
\textbf{Image} & \textbf{Captions} \\
\midrule
    \begin{minipage}[c]{\linewidth}
        \centering
        \includegraphics[width=0.8\linewidth]{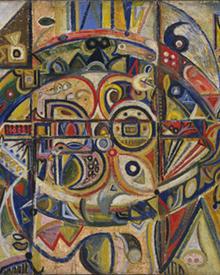} \\[0.5ex]
        Desert \\ Richard Pousette Dart
        \vspace{0.25em}
    \end{minipage}
    &
    \begin{minipage}[c]{\linewidth}
        The artwork is a vibrant and abstract composition that appears to be a painting. It features a dense arrangement of geometric shapes, lines, and dots in a variety of colors, including red, blue, yellow, and black. The shapes and lines intersect and overlap, creating a sense of complexity and depth. The mood or emotion conveyed by the artwork is one of chaos and complexity, with the abstract nature of the shapes and lines leaving room for interpretation. The key artistic techniques and style evident in the painting are reminiscent of the Cubist style, which is characterized by the use of geometric shapes and lines to create a sense of depth and dimensionality. The use of bold, contrasting colors also adds to the overall visual impact of the artwork. The painting could be associated with the Cubist period or school, which was active in the early 20th century. This period was known for its emphasis on abstract forms and the use of geometric shapes to represent the three-dimensional world. The artwork's abstract nature and the use of geometric shapes and lines suggest that it may be influenced by or a part of the Cubist movement.
    \end{minipage}
    \\

    \cmidrule(lr){1-2}

    \begin{minipage}[c]{\linewidth}
        \centering
        \includegraphics[width=0.8\linewidth]{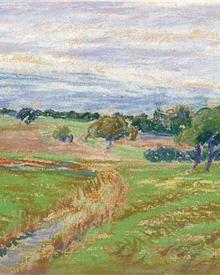} \\[0.5ex]
        The Hills of Thierceville \\ Camille Pissarro
        \vspace{0.25em}
    \end{minipage}
    &
    \begin{minipage}[c]{\linewidth}
        The artwork depicts a serene rural landscape. The scene is dominated by a field of tall grass, with a winding path cutting through it. The path leads the viewer's eye towards a line of trees in the distance, suggesting a sense of depth and perspective. The sky above is filled with clouds, adding a sense of tranquility and calmness to the scene. The mood or emotion conveyed by the artwork is one of peace and solitude. The natural elements, such as the grass, trees, and sky, are rendered with a loose, impressionistic style, characterized by visible brushstrokes and a focus on the interplay of light and color. This style is often associated with the Impressionist period, which sought to capture the fleeting effects of light and color in the natural world. The artwork's subject matter, the rural landscape, is a common theme in many art movements, including the Impressionists, who often sought to capture the beauty of the natural world. The loose, impressionistic style, combined with the tranquil subject matter, suggests that this artwork could be a representation of the Impressionist style, particularly if it was created during the late 19th or early 20th century.
    \end{minipage}
    \\

\bottomrule
\end{tabularx}
\label{tab:image_capts_examples}
\end{table*}

Examples of generated captions for both images and audio are reported in Tables~\ref{tab:image_capts_examples} and~\ref{tab:audio_capts_examples}, respectively. Notably, the level of detail in image captions is much higher than that of audio captions. This gap may be due to the choice of captioning models; a similar difference was observed with other models, as shown in Appendices~\ref{abstud:image_capt}–\ref{abstud:audio_capt}.

\begin{table*}[t]
\centering
\caption{Examples of $\ArtSound$ audio captions generated using LP-MusicCaps + Qwen3.}
\footnotesize
\renewcommand{\arraystretch}{1.1}
\begin{tabularx}{0.95\textwidth}{@{} >{\bfseries}M{2.4cm} | X @{}}
\toprule
\textbf{Audio} & \textbf{Captions} \\
\midrule
    \begin{minipage}[c]{\linewidth}
        \centering
        Audio \#1 \\ \texttt{(001/001556)}
    \end{minipage} &
    \begin{minipage}[c]{\linewidth}
        This is a music piece that features a low quality recording with an ambient song consisting of sustained synth pad chords, sustained synth bass, mellow piano chords, and a short plucked melody. It sounds calming, relaxing, and hypnotic—like something you would hear in yoga sessions or a spa center.
    \end{minipage} \\

    \cmidrule(lr){1-2}

    \begin{minipage}[c]{\linewidth}
        \centering
        Audio \#2 \\ \texttt{(001/001140)}
    \end{minipage} &
    \begin{minipage}[c]{\linewidth}
        This is a music piece that features a male voice singing the main melody, accompanied by percussion playing a simple beat in common time, with the bass playing the root notes of the chords and a piano playing arpeggiated chords in the background. The mood is romantic, and it could be played in a romantic movie. Additionally, it has elements of an instrumental rock music piece recorded in the style of blues, with a male vocalist singing melodically in the lead, the main tune played by a piano, and the electric guitar and bass guitar playing in the background. The rhythmic background consists of a slow tempo acoustic drum beat, and the atmosphere is melancholic, making it suitable for the soundtrack of a drama TV series, especially during scenes where a character is trying to break free.
    \end{minipage} \\
    
    \bottomrule
\end{tabularx}
\label{tab:audio_capts_examples}
\end{table*}

Additional experiments were conducted for both image and audio captioning using different models, as well as for caption fusion in audio captioning. For image captioning, some models~\citep{castellano2023exploring, bai2021explain} were initially considered but later discarded because they were outdated, while others~\citep{bai2025qwen2, xu2025qwen2} produced either overly long and unbalanced descriptions or generic captions without support for prompt conditioning~\citep{li2023lavis}. Examples are provided in Appendix~\ref{abstud:image_capt}. For audio captioning, several models~\citep{liu2026mumu, deng2024musilingo, chu2024qwen2, xu2025qwen2} were identified. However, \citep{liu2026mumu} was excluded due to high VRAM requirements and low-quality captions. The remaining models~\citep{deng2024musilingo, chu2024qwen2, xu2025qwen2} generated generally accurate but overly detailed captions, often repeating the same concepts, and occasionally included questions to the user or even fabricated responses within the caption itself, making the outputs unsuitable. Examples of such captions can be found in Appendix~\ref{abstud:audio_capt}. For audio caption fusion, we also evaluated smaller LLMs (LLaMa3.1-8B-Instruct~\citep{grattafiori2024llama} and Qwen3-8B~\citep{yang2025qwen3}) as alternatives to Qwen3-14B. Using the $\ACscore$ (Equation~\ref{eq:aud_capt_score}), Qwen3-14B performed better, demonstrating stronger preservation of both meaning and original wording. Table~\ref{tab:audio_capt_llms} reports these results, including $\ACscore$ and the underlying metrics, ROUGE-1 and BERT-Score, for completeness.

\begin{table}[t]
  \centering
  \caption{Average metrics computed during the LLM evaluation for audio caption fusion. Metrics include ROUGE-1, BERT-Score, and the custom $\ACscore$. Best scores are in \textbf{bold}, while \underline{underlined} values indicate those that are close to the best.}
    \begin{tabular}{lccc}
      \toprule
      \textbf{LLM} & \textbf{ROUGE-1} & \textbf{BERT-Score} & $\ACscore$ \\
      \midrule
      LLaMa3.1-8B-Instruct & 0.5373 & 0.9056 & 0.7951 \\
      Qwen3-8B & \underline{0.6301} & \underline{0.9260} & \underline{0.8372} \\
      Qwen3-14B & \textbf{0.6870} & \textbf{0.9312} & \textbf{0.8579} \\
      \bottomrule
    \end{tabular}
  \label{tab:audio_capt_llms}
\end{table}


\subsection{Artwork–Music Pairing} 
\label{sec:pairing_approacches}

Given the captioned images and audio collections, our goal was to construct semantically meaningful image–audio pairs. To this end, we evaluated several embedding-based similarity approaches using representations from ImageBind~\citep{girdhar2023imagebind}.

Since ImageBind embeds representations from multiple modalities (e.g., text, images, and audio) into a shared representation space, this enables direct comparison across all three modalities. Indeed, we evaluated several pairing strategies by comparing embeddings across these different modalities.

Pairing was based on the cosine similarity of embeddings from a data source, which can be either an image or an audio file, in a specific modality, either raw data or captions. The pairing process began with items that exhibit the highest similarity. We specifically focused on audio items, pairing each with the images most similar to it.

This procedure can be formalized as follows. Let $\mathcal{A} = \{a_1, \ldots, a_N\}$ denote the set of audio items and $\mathcal{I} = \{i_1, \ldots, i_M\}$ the set of images. Each item $x \in \mathcal{A} \cup \mathcal{I}$ is embedded into a shared space of dimensionality $d$, producing an embedding $e^{(v)}_x \in \mathbb{R}^d$, where $v \in \{\text{raw}, \text{caption}\}$ indicates the modality used. The cosine similarity between an image $i$ and an audio item $a$, under modalities $v_i$ and $v_a$, is defined as:
\begin{equation}
sim(i, a;v_{\text{i}}, v_{\text{a}}) =
\frac{e^{(v_i)}_{\text{i}}\cdot e^{(v_a)}_{\text{a}}}{\|e^{(v_i)}_{\text{i}}\|\,\|e^{(v_a)}_{\text{a}}\|}.
\label{eq:cos_sim_embeds}
\end{equation}
For each audio item $a_j \in \mathcal{A}$, pairing selects the image:
\begin{equation}
p^*_j= \underset{i \in \mathcal{I}_{\text{avail}}}{\text{argmax}} \ sim(i_{\text{}}, a_{\text{j}};v_{\text{i}}, v_{\text{a}}),
\label{eq:pair_computation}
\end{equation}
where $\mathcal{I}_{\text{avail}}$ is the current set of unpaired images, and the selected image $p^*_j$ is subsequently removed from $\mathcal{I}_{\text{avail}}$. The process continues until all $a_N \in \mathcal{A}$ ($N = 105{,}884$) are paired. A limitation of this greedy strategy arises toward the end of the pairing process: after matching the items with the highest similarity values, the remaining items must be paired despite their low similarity scores. To enable downstream analysis of this effect, we included the computed similarity values in the final dataset.

\begin{figure*}[t]
    \centering
    \begin{minipage}{0.48\linewidth}
        \centering
        \subfloat[Similarity distribution (Image–-Audio)]{
            \includegraphics[width=\linewidth]{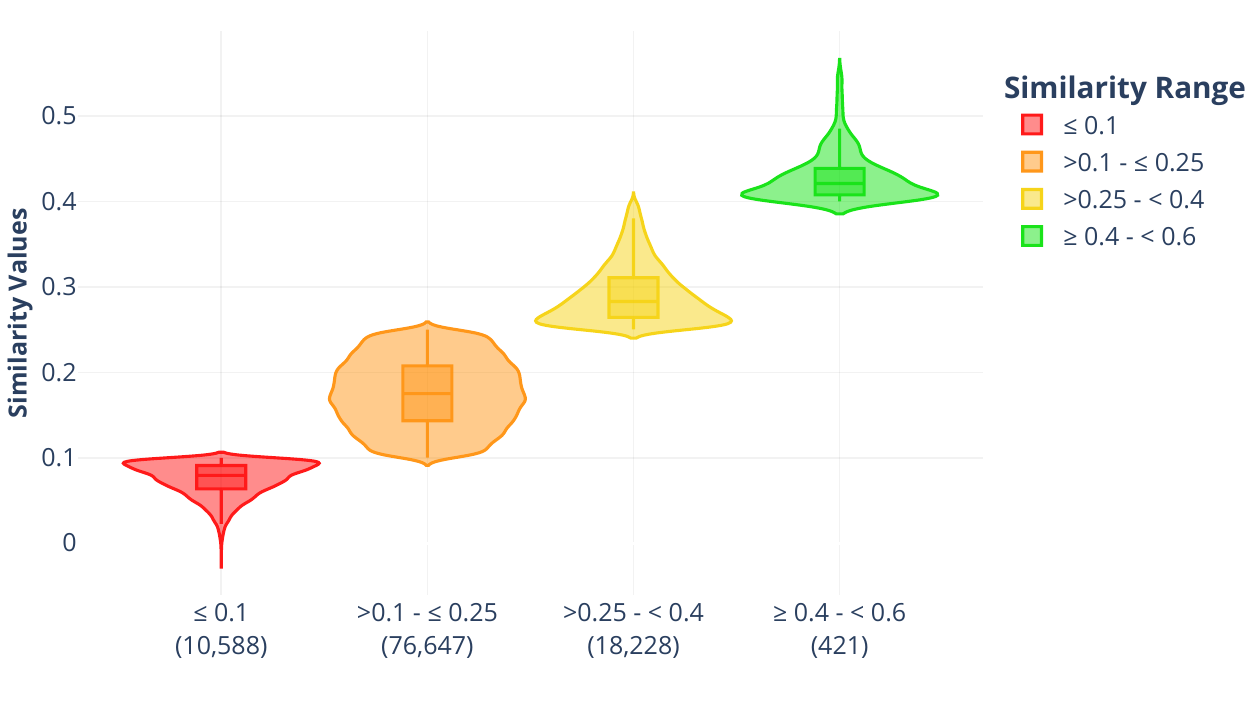}
            \label{fig:img_aud_imgbind_violin}
        }
    \end{minipage}
    \hfill
    \begin{minipage}{0.48\linewidth}
        \centering
        \subfloat[Similarity distribution (Image–-Audio Caption)]{
            \includegraphics[width=\linewidth]{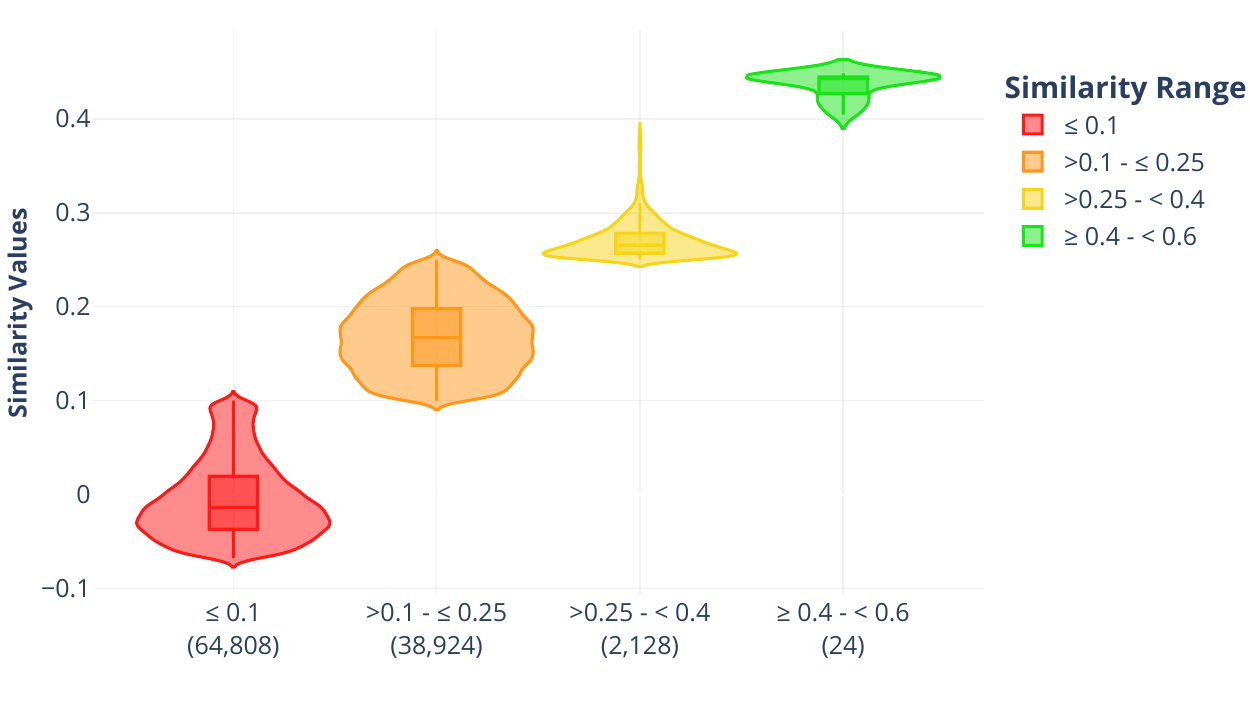}
            \label{fig:img_audc_imgbind_violin}
        }
    \end{minipage}
    \hfill
    \begin{minipage}{0.48\linewidth}
        \centering
        \subfloat[Similarity distribution (Image Caption–-Audio)]{
            \includegraphics[width=\linewidth]{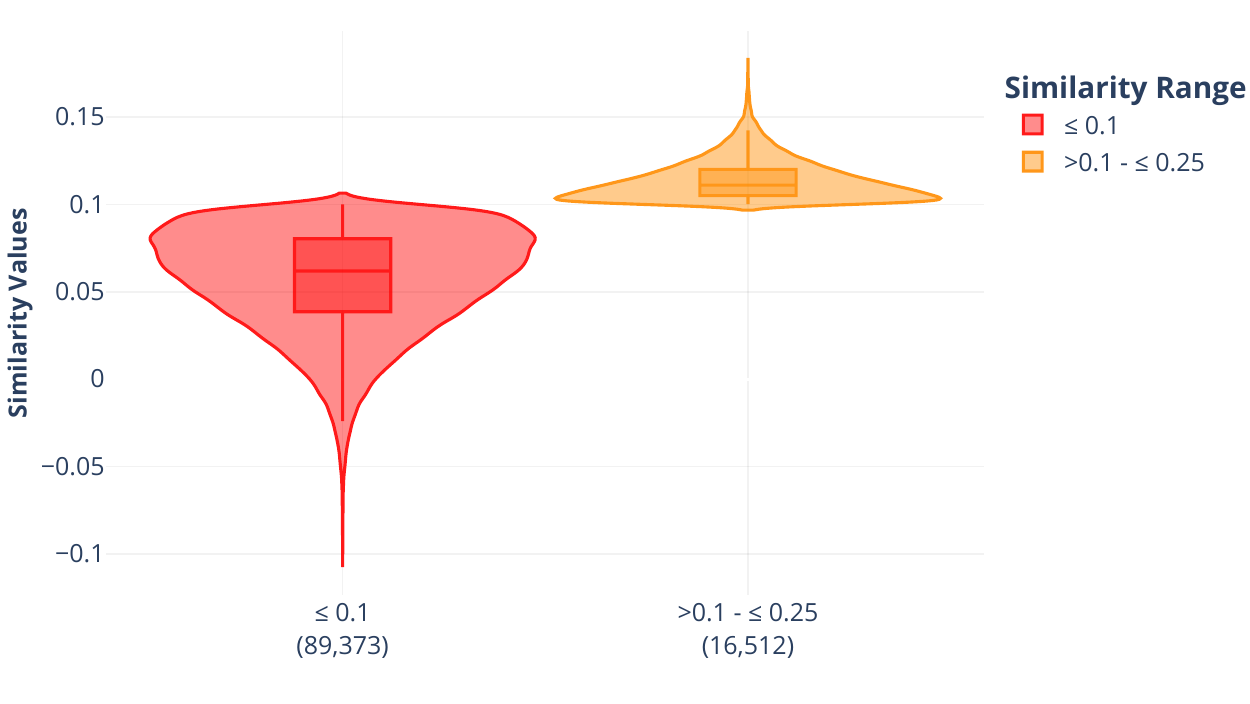}
            \label{fig:imgc_aud_imgbind_violin}
        }
    \end{minipage}
    \hfill
    \begin{minipage}{0.48\linewidth}
        \centering
        \subfloat[Similarity distribution (Image Caption–-Audio Caption)]{
            \includegraphics[width=\linewidth]{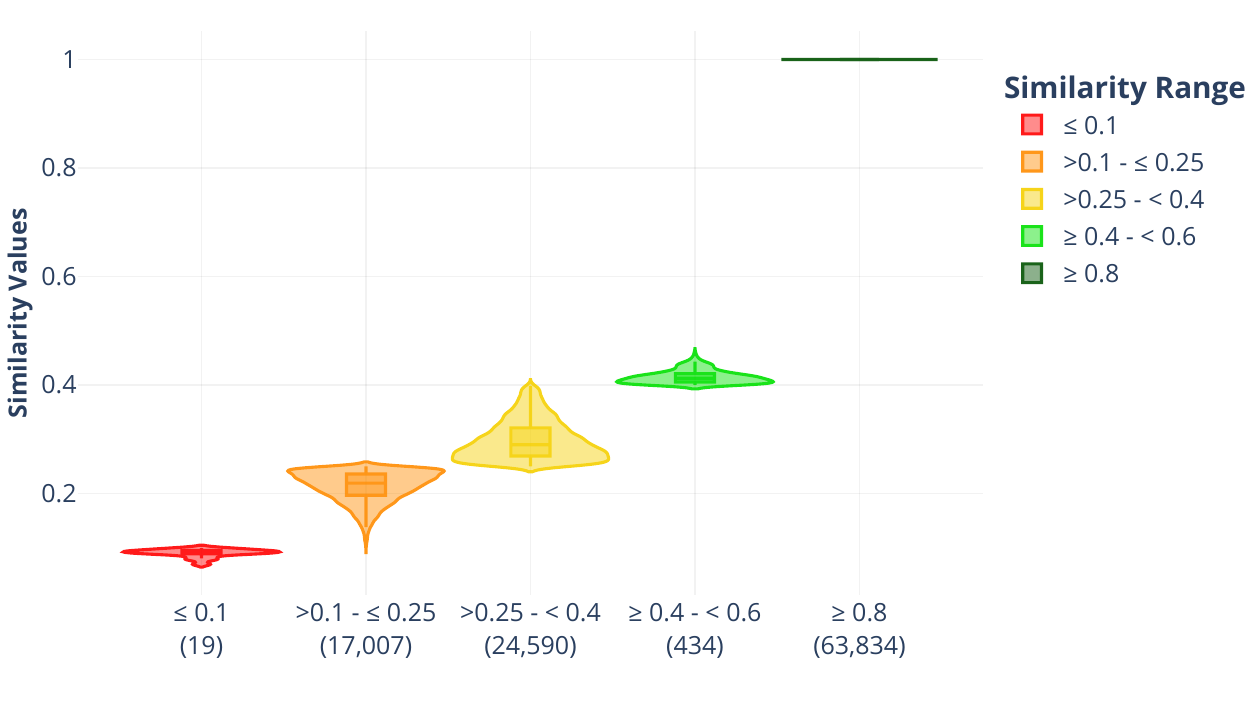}
            \label{fig:imgc_audc_imgbind_violin}
        }
    \end{minipage}
    \caption{Violin plots of similarity score distributions for image–audio pairing strategies. Scores, computed from combinations of image, audio, and caption embeddings using ImageBind, are grouped into low ($\le$0.25), medium (0.25–0.6), and high ($\ge$0.6) ranges, highlighting which model–modality pairs produce stronger or weaker alignments.}
    \Description{This figure shows violin plots of similarity scores for different image–audio pairing strategies, illustrating how combinations of image, audio, and caption embeddings from ImageBind distribute across low, medium, and high alignment ranges. Highlights which model–modality pairs achieve stronger semantic alignment.}
    \label{fig:violin_plots_imagebind}
\end{figure*}

We analyzed the computed similarities to assess potential modality gaps and to investigate whether comparisons within the same modality (e.g., Text–Text) produced higher similarity scores than cross-modal comparisons (e.g., Image–Text or Audio–Text). Objective results for each pairing approach are reported in Table~\ref{tab:pairing_approaches_results}, and Figure~\ref{fig:violin_plots_imagebind} illustrates the distribution of similarity values across fixed ranges, providing a clear view of how many values fall in lower versus higher regions.

\begin{table}[b]
    \centering
    \caption{Summary of similarity statistics (minimum, maximum, mean, and standard deviation) for the pairing approaches across images, audio, image captions, and audio captions. All approaches use ImageBind embeddings for the employed modalities. \textbf{Bold} numbers denote the highest values, while \underline{underlined} numbers mark the second highest values.}
    \begin{tabular}{l|cccc|cccc}
        \toprule
        \multirow{2}{*}{\textbf{Pairing Approach}} 
        & \multirow{2}{*}{\textbf{Image}} 
        & \multirow{2}{*}{\begin{tabular}[t]{@{}c@{}}\textbf{Image}\\\textbf{Caption}\end{tabular}}
        & \multirow{2}{*}{\textbf{Audio}} 
        & \multirow{2}{*}{\begin{tabular}[t]{@{}c@{}}\textbf{Audio}\\\textbf{Caption}\end{tabular}} 
        & \multicolumn{4}{c}{\textbf{Similarity Values}}  \\
        & & & & & \textbf{Min}     & \textbf{Max}    & \textbf{Avg}    & \textbf{Std Dev}  \\
        \midrule  
        Image--Audio    & \ding{51} & --- & \ding{51} & --- & \underline{$-$0.0216} & \underline{0.5518} & \underline{0.1864} & \underline{0.0697}    \\
        Image--Audio Caption   & \ding{51} & --- & --- & \ding{51}& $-$0.0676 & 0.4482 & 0.0644 & 0.0970    \\
        Image Caption--Audio   & --- & \ding{51} & \ding{51} & ---& $-$0.1011 & 0.1794 & 0.0658 & \textbf{0.0344}    \\
        Image Caption--Audio Caption  & --- & \ding{51} & --- & \ding{51}& \textbf{0.0699}  & \textbf{1.0000} & \textbf{0.7081} & 0.3613    \\
        \bottomrule
    \end{tabular}
    \label{tab:pairing_approaches_results}
\end{table}

Overall, the evaluated pairing strategies yielded relatively low similarity scores (mean values ranging from 0.0644 to 0.7081), indicating a substantial modality gap between images and audio. The strongest alignment was observed when pairing image captions with audio captions, which achieved the highest mean similarity (0.7081), suggesting that textual representations provide a more consistent shared space than raw modalities. In contrast, the combination of raw images and audio captions yielded the lowest performance, highlighting the difficulty of directly bridging visual and auditory information through mismatched representations.


Finally, a qualitative inspection of the computed pairs led us to select the Image--Audio approach as the best one. Although its average similarity score was not the highest, the paired items appeared more semantically aligned than those produced by the other methods. The pairs obtained through this approach were therefore used to build $\ArtSound$, our final dataset comprising 105{,}884 artwork-music pairs, each enriched with high-quality image and audio captions.

We also evaluated alternative embedding models, including CLIP~\citep{radford2021learning}, CLAP~\citep{elizalde2023clap}, and two text encoders (RoBERTa~\citep{liu2019roberta} and Modern-BERT~\citep{warner2025smarter}), but qualitative analysis revealed that these models frequently produced unrelated image-audio pairs, leading us to discard them. Additionally, we also tested other pairing approaches, described in Appendix~\ref{abstud:pairing_approaches}, including pairing artworks with audio spectrograms and audio-generated images using a small subset of $\ArtGraph$ artworks and music tracks from the Free Music Archive dataset. Although overall results were poor, some generated pairs indicated that the artwork style and music genre shared similar ideals. Further details are provided in Appendix~\ref{abstud:pairing_approaches}.

\section{$\ArtToMus$}
\label{sec:methodology}

\begin{figure*}[b]
  \centering
  \includegraphics[width=.85\linewidth]{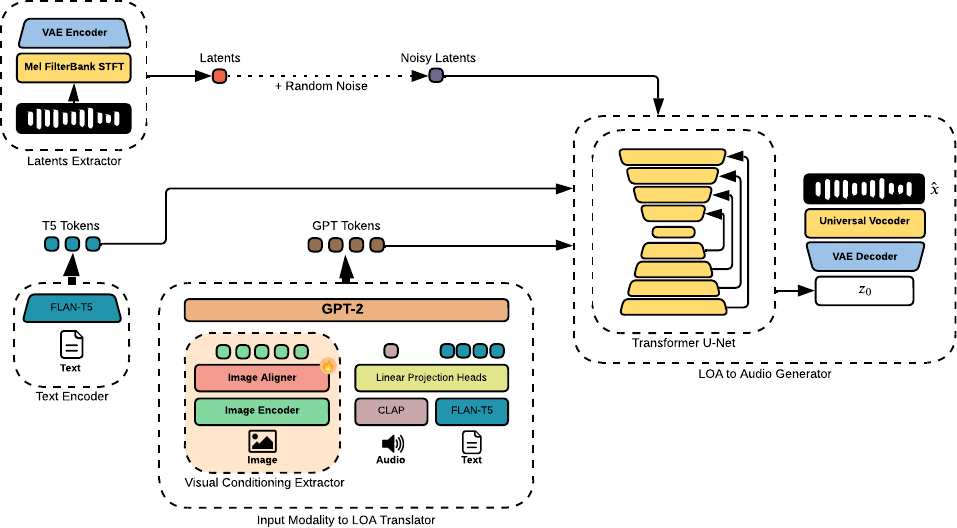}
    \caption{Overview of the $\ArtToMus$ architecture. Inspired by the generative paradigm of AudioLDM~2, $\ArtToMus$ reformulates the conditioning interface to enable direct artwork-to-music generation. A \textit{Visual Conditioning Extractor} and an \textit{Image Aligner} learn to project image embeddings into GPT-2's LoA embedding space, establishing cross-modal alignment without textual intermediates. The pretrained AudioLDM~2 backbone is used as a frozen generative prior. GPT-2 operates as a modality-agnostic LoA translator, integrating visual conditioning within the latent diffusion process. Music is encoded into mel-spectrogram latents by the \textit{Latent Extractor} and synthesized into audio by the \textit{LoA-to-Audio Generator}.}
  \Description{The figure illustrates ArtToMus architecture, explaining how visual information is injected into a pretrained text-to-audio diffusion backbone with minimal architectural changes. It highlights the role of GPT-2 as a shared intermediate representation that aligns visual, textual, and audio modalities while keeping most components frozen.}
  \label{fig:art2musreimagined_architecture}
\end{figure*}

This section presents the architecture of $\ArtToMus$ (Figure~\ref{fig:art2musreimagined_architecture}), a multimodal framework inspired by the generative paradigm of AudioLDM~2~\citep{liu2024audioldm}, but reformulated to address a fundamentally different problem: direct artwork-to-music generation. While AudioLDM~2 operates under a text-to-audio paradigm, $\ArtToMus$ is designed around a visual-to-audio formulation that intentionally removes linguistic supervision. Rather than extending the original task, we adopt its latent diffusion backbone as a pretrained generative prior and redesign the conditioning interface to support direct cross-modal alignment from visual representations. The core contribution lies not in modifying the diffusion architecture itself, but in redefining how conditioning information is represented and aligned. In particular, we introduce a visual-to-LoA (Language of Audio) interface that learns to encode image embeddings into the latent token space originally developed for audio semantics, enabling the diffusion model to operate under a new cross-modal conditioning regime.

\subsection{Base Architecture}

AudioLDM~2~\citep{liu2024audioldm} applies latent diffusion modeling~\citep{rombach2022high} to audio generation. Audio signals are first converted into mel-spectrograms and encoded into latent vectors $z_0$ using a Variational Autoencoder (VAE). A diffusion process gradually perturbs $z_0$ into $z_T \sim \mathcal{N}(0, 1)$, and a UNet-based denoiser $\epsilon_\theta$ is trained to invert this process. The objective is
\begin{equation} \label{eq:ldm} \mathcal{L}_{\text{LDM}} = \mathbb{E}_{z_0, \epsilon \sim \mathcal{N}(0,1), t} \left[ \left\| \epsilon - \epsilon_\theta(z_t, t, c) \right\|_2^2 \right], \end{equation}
where $c$ denotes the conditioning information, and $t$ is the diffusion timestep sampled from $\{1, \ldots , T\}$. Unlike text-to-image diffusion models, AudioLDM~2 encodes conditioning information for generation in the LoA embedding space—a continuous representation of audio signals that captures both semantic and acoustic information. This space is constructed using AudioMAE~\citep{huang2022masked}, a self-supervised masked autoencoder trained to reconstruct masked spectrogram patches and learn meaningful audio representations.

For text-to-audio generation, a GPT-2 model is fine-tuned to map textual prompts directly into the LoA space. This formulation leads to a simpler objective, as the diffusion model is conditioned on LoA embeddings rather than directly on text representations. During training, the conditioning signal alternates between ground-truth LoA features and GPT-2–predicted LoA features, promoting cross-modal generalization while preserving the semantic richness of the LoA embeddings.

The denoiser $\epsilon_\theta$ is a T-UNet (Transformer UNet) similar to those used in image generation~\citep{rombach2022high}. Convolutional downsampling and upsampling layers are interleaved with transformer blocks containing both self-attention and cross-attention, allowing the model to incorporate LoA features into the denoising process. The denoised latent representation is decoded back into a mel-spectrogram by the VAE decoder, and a pretrained HiFiGAN vocoder~\citep{kong2020hifi} converts the spectrogram into the final audio waveform.

\begin{figure*}[!t]
    \centering
    \begin{minipage}{0.45\linewidth}
        \centering
        \subfloat[\textit{Visual Conditioning Extractor}. A pretrained image encoder (e.g., CLIP or ImageBind) produces an embedding that is subsequently mapped to GPT-2’s token space (dimension 768).]{
            \includegraphics[width=.45\linewidth]{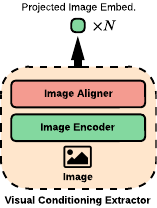}
            \label{fig:visual_conditioning_extractor}
        }
    \end{minipage}
    \hfill
    \begin{minipage}{0.45\linewidth}
        \centering
        \subfloat[\textit{Image Aligner}. If the input embedding is 512-dimensional (CLIP), it is upscaled to 1024 dimensions; otherwise (ImageBind) it bypasses this step. A projection layer then maps all embeddings to $N$ tokens in GPT-2’s 768-dimensional space.]{
            \includegraphics[width=.85\linewidth]{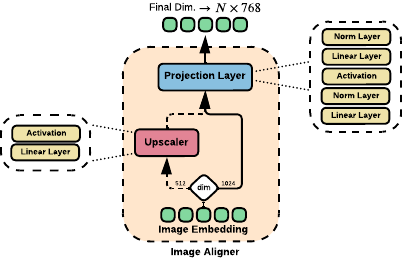}
            \label{fig:image_aligner}
        }
    \end{minipage}

    \caption{The \textit{Visual Conditioning Extractor} and \textit{Image Aligner} modules.}
    \Description{The figure focuses on the visual conditioning pathway, showing how image embeddings are normalized and projected into a common token-based representation. It clarifies how different vision encoders are handled to ensure compatibility with GPT-2.}
    \label{fig:visual_conditioning_extractor_and_aligner}
\end{figure*}

\subsection{Visual Conditioning Extractor}

AudioLDM~2 was originally designed to handle only textual conditioning, meaning its internal LoA representation lacks a native mechanism for incorporating visual information. Simply fine-tuning the model with image–audio pairs does not enable cross-modal alignment, because the denoiser and GPT-2 expect conditioning expressed in a tokenized text space. To overcome this architectural limitation, $\ArtToMus$ introduces the \textit{Visual Conditioning Extractor}, a new component that establishes a dedicated path for visual semantics to enter the LoA space (Figure~\ref{fig:visual_conditioning_extractor}). 

Given an artwork image, a pretrained vision encoder extracts an embedding of dimension 512 (CLIP) or 1024 (ImageBind). These embeddings are then passed to the \textit{Image Aligner} (Figure~\ref{fig:image_aligner}), composed of:
\begin{itemize}
    \item Upscaler: expands 512-dimensional CLIP embeddings to 1024 dimensions;
    \item Projection Layer: maps each 1024-dimensional embedding to $N$ tokens in GPT-2’s 768-dimensional embedding space.
\end{itemize}

This modular design ensures compatibility with heterogeneous visual encoders and provides a stable intermediate representation. Empirically, using a 1024-dimensional staging space before projection yielded more faithful, stylistically coherent music than directly projecting lower-dimensional embeddings.




\subsection{Music Generation Process}

Given an artwork, the \textit{Visual Conditioning Extractor} produces image-derived conditioning tokens, which GPT-2 transforms into LoA embeddings $c_{\text{vis}}$. In parallel, the textual prompt \textit{``Music representing the content of this artwork''} is encoded using FLAN-T5~\citep{chung2024scaling}, producing $c_{\text{text}}$. The final conditioning vector $c$ combines both modalities.

At inference time, generation starts from Gaussian noise $z_T \sim \mathcal{N}(0, 1)$ and proceeds through the reverse diffusion process. To improve efficiency and enable deterministic sampling, we adopt a DDIM scheduler \citep{song2021denoising}, enabling high-quality generation with fewer denoising steps. Let $\bar{\alpha}_t = \prod_{s=1}^{t} \alpha_s$ denote the cumulative product of the variance schedule coefficients, where each $\alpha_t = 1 - \beta_t$ controls the amount of signal preserved at timestep $t$, and $\beta_t$ is the predefined noise variance. The quantity $1 - \bar{\alpha}_t$ therefore represents the accumulated noise level. At each timestep $t$, the model predicts the noise component $\epsilon_\theta(z_t, t, c)$ and computes an estimate of the clean latent:
\begin{equation}
\hat{z}_0
=
\frac{1}{\sqrt{\bar{\alpha}_t}}
\left(
z_t
-
\sqrt{1-\bar{\alpha}_t}\,
\epsilon_\theta(z_t, t, c)
\right).
\end{equation}
The DDIM update rule then deterministically maps $z_t$ to $z_{t-1}$ as
\begin{equation}
z_{t-1}
=
\sqrt{\bar{\alpha}_{t-1}}\,\hat{z}_0
+
\sqrt{1-\bar{\alpha}_{t-1}}\,
\epsilon_\theta(z_t, t, c).
\end{equation}
After iterating down to $t=0$, the final latent $\hat{z}_0$ is decoded into a mel-spectrogram and converted into a waveform via the VAE decoder and neural vocoder.

\subsection{Training Algorithm}

Training $\ArtToMus$ requires learning a cross-modal alignment that is fundamentally absent in AudioLDM~2: the model must associate fine-grained visual embeddings with the structure of music encoded in the LoA space. This constitutes a new learning problem, since AudioLDM~2 is never exposed to images and lacks any prior for relating visual semantics to musical attributes.

Specifically, training relies on triplets composed of an artwork, its paired music track, and a fixed guiding prompt, which is the same one used at inference time (i.e., \textit{``Music representing the content of this artwork''}). Classifier-free guidance is enabled by also providing a negative prompt (e.g., \textit{``Low quality''}), optionally augmented to discourage unwanted genres. The negative prompt helps the model steer the generation away from undesired aspects, improving overall quality and ensuring the output aligns more closely with the desired intent. Each paired music track is converted into a mel-spectrogram and encoded into a clean latent vector $z_0$ using the AudioLDM~2 VAE. A timestep $t$ is sampled uniformly, and noise is added to obtain $z_t$. The diffusion model is trained to predict the added noise following Equation~\eqref{eq:ldm}, conditioned on the visual LoA tokens and auxiliary textual embeddings.

Crucially, only the parameters of the \textit{Image Aligner} are updated. All AudioLDM~2 components, including the denoiser and the GPT-2 LoA translator, remain frozen. This preserves the pretrained generative prior while enabling the system to learn a robust mapping from artworks to LoA conditioning features. Through this reformulation, $\ArtToMus$ demonstrates that a text-conditioned diffusion prior can be repurposed for direct visual-to-music generation via representation alignment, without altering the underlying generative architecture.

In practice, depending on whether ImageBind or CLIP was used as the visual encoder, either only the \textit{Projection Layer} or both the \textit{Upscaler} and \textit{Projection Layer} were trained. Training follows the standard DDPM objective, where the model predicts the noise added to the clean latent $z_0$. The forward diffusion process is defined as
\begin{equation}
z_t
=
\sqrt{\bar{\alpha}_t}\, z_0
+
\sqrt{1-\bar{\alpha}_t}\, \epsilon,
\qquad
\epsilon \sim \mathcal{N}(0,I),
\end{equation}
where $\bar{\alpha}_t = \prod_{s=1}^{t} (1-\beta_s)$ is the cumulative product of the variance schedule coefficients, and $1-\bar{\alpha}_t$ represents the accumulated noise level. For a given timestep $t$, the base noise-prediction loss is
\begin{equation}
\mathcal{L}_{\text{MSE}}(t)
=
\left\|
\epsilon - \epsilon_\theta(z_t, t, c)
\right\|_2^2.
\end{equation}
The signal-to-noise ratio at timestep $t$ is therefore
\begin{equation}
\mathrm{SNR}_t
=
\frac{\bar{\alpha}_t}{1-\bar{\alpha}_t},
\end{equation}
which directly reflects the relative strength of the preserved signal $\sqrt{\bar{\alpha}_t}$ and the injected noise $\sqrt{1-\bar{\alpha}_t}$.
Following~\citep{hang2023efficient}, the timestep loss is reweighted as
\begin{equation}
\mathcal{L}_t
=
\frac{\min(\mathrm{SNR}_t, \gamma)}{\mathrm{SNR}_t}
\cdot
\left\|
\epsilon - \epsilon_\theta(z_t, t, c)
\right\|_2^2,
\end{equation}
with $\gamma = 5.0$.

This formulation preserves the original noise-prediction objective while modulating each timestep’s contribution according to its SNR. Timesteps dominated by noise (small $\bar{\alpha}_t$) are downweighted, whereas those with stronger signal content receive relatively higher emphasis, leading to more stable optimization and improved sample quality.

\section{Experiments}
\label{sec:experiments}

This section describes the experimental protocol adopted to train and evaluate $\ArtToMus$. We first provide an exploratory analysis of the $\ArtSound$ dataset, then detail the training setup, performance metrics, and model configurations. Objective and subjective evaluation results are subsequently reported.

\subsection{Exploratory Data Analysis}

We conducted a preliminary analysis of $\ArtSound$ to assess the distribution of artwork–music similarities and explore potential associations between artistic styles and music genres. Artwork–music pairs were constructed using ImageBind embeddings (Section~\ref{sec:pairing_approacches}), selecting for each music track the artwork most similar to it.

Similarity values, computed as the cosine similarity between artwork (image) and music (audio) embeddings extracted using ImageBind, range from $-0.0216$ to $0.5518$, with an average of $0.1864$ (Table~\ref{tab:pairing_approaches_results}). Although similarities are generally modest—consistent with the cross-modal nature of the task—more than $50{,}000$ pairs exceed the mean value, indicating a smooth and continuous distribution of similarity scores rather than a sharp separation between low and high values. A small manual inspection (30 samples) confirmed that many pairs appeared intuitively meaningful, although others remained ambiguous, reflecting the inherently subjective nature of cross-modal alignment.

\begin{figure*}[t]
    \centering
    \includegraphics[width=1.\textwidth]{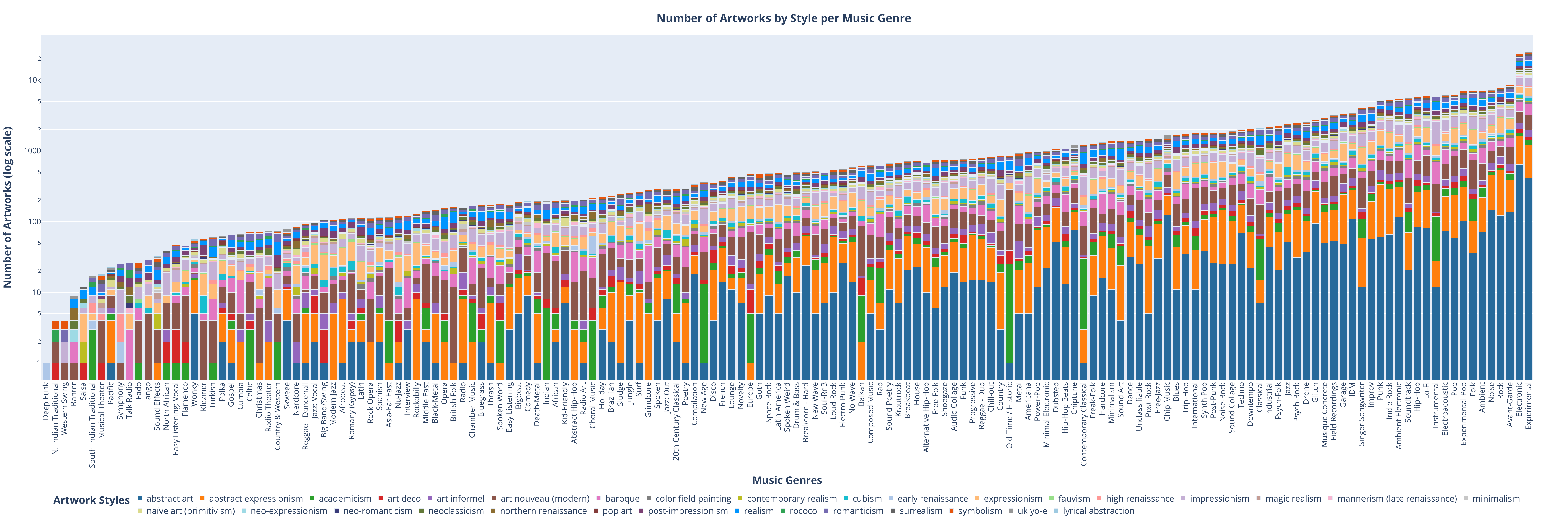}
    \caption{Distribution of artworks by music genre, segmented by artistic style. 
    \textbf{\textcolor{abstractart_color}{Abstract Art}}, 
    \textbf{\textcolor{abst_expressionism_color}{Abstract Expressionism}},
    \textbf{\textcolor{artnouveau_color}{Art Nouveau (Modern)}}, and
    \textbf{\textcolor{expressionism_color}{Expressionism}}
    appear prominently across most genres.
    Conversely, 
    \textbf{\textcolor{earlyreinass_color}{Early Renaissance}} and
    \textbf{\textcolor{symbolism_color}{Symbolism}}
    are less frequent.}
    \label{fig:artwstyle_by_musicgenre_fulldata}
    \Description{This figure illustrates the relationship between artistic styles and associated music genres, revealing dominant and underrepresented styles across the dataset. It helps contextualize stylistic biases and coverage in the artwork–music pairing.}
\end{figure*}

We next analyzed artwork–music associations by examining the distribution of 32 artistic styles across 161 music genres (Figure~\ref{fig:artwstyle_by_musicgenre_fulldata}). Several broad trends emerged: (\textit{i}) genres such as Experimental, Electronic, and Avant-Garde tend to co-occur with modern and perceptually driven movements (e.g., Impressionism, Expressionism); (\textit{ii}) less represented genres (e.g., Deep Funk, Western Swing) show a more heterogeneous style distribution; and (\textit{iii}) popular genres (Hip-Hop, Jazz, Blues) often align with styles emphasizing emotional or narrative expressiveness. These trends are not intended as definitive aesthetic correspondences, but they highlight that embedding-based pairing captures weak yet interpretable cross-modal regularities.

\subsection{Experimental Setting}

The $\ArtSound$ dataset was split into $\mathcal{D}_\text{train}$ and $\mathcal{D}_\text{test}$ using a 90/10 stratified split over artistic styles. A validation set $\mathcal{D}_\text{val}$ of 100 artworks was sampled from $\mathcal{D}_\text{test}$. During validation, the model generated 10-second samples using 200 inference steps and the negative prompt \textit{``Low quality''}.

Training was run for 50 epochs, and used AdamW with default hyperparameters, a learning rate of $2{\times}10^{-5}$, 1{,}324 warmup steps, 37{,}200 total steps, batch size 128 (32 per GPU across 4$\times$A100 GPUs), and 1 gradient accumulation step. Only the image-to-LoA interface was updated; all AudioLDM~2 components remained frozen. The full 50-epoch training run took approximately 60 hours.

\subsection{Performance Metrics}

Caption quality was evaluated using $\ICscore$ and $\ACscore$ (Section~\ref{sec:materials}). Generated audio was assessed using three complementary metrics.

KL-Divergence (KL-Div) measures how closely the feature distribution of the generated music matches that of the ground-truth audio. Given distributions $\mathcal{P}$ (ground truth) and $\mathcal{Q}$ (generated), lower values indicate better alignment:
\begin{equation}
\text{KL-Div}(\mathcal{P}\|\mathcal{Q}) = 
\sum_i \mathcal{P}(i)\log\frac{\mathcal{P}(i)}{\mathcal{Q}(i)}.
\end{equation}

Fréchet Audio Distance (FAD)~\citep{kilgour2018fr} evaluates perceptual realism by comparing VGGish embeddings of generated and reference audio, modeled as Gaussians:
\begin{equation}
\text{FAD} = \|\mu_b - \mu_e\|^2 +
\mathrm{Tr}\!\left(\Sigma_b + \Sigma_e - 2(\Sigma_b\Sigma_e)^{1/2}\right),
\end{equation}
where $(\mu_b,\Sigma_b)$ and $(\mu_e,\Sigma_e)$ are the mean and covariance of reference and generated embeddings. Lower scores indicate closer perceptual similarity to real audio.

ImageBind Score (IBSc) measures cross-modal consistency by computing the cosine similarity between ImageBind embeddings of the artwork and the generated audio:
\begin{equation}
\text{IBSc}(e_{\text{art}}, e_{\text{gen}}) =
\frac{e_{\text{art}}\cdot e_{\text{gen}}}{\|e_{\text{art}}\|\,\|e_{\text{gen}}\|}.
\end{equation}
We report IBSc both with respect to the artwork and to the paired ground-truth track; higher values indicate stronger visual–audio semantic alignment.

Together, these metrics quantify distributional similarity (KL-Div), perceptual realism (FAD), and cross-modal coherence (IBSc).

\subsection{Experimental Results}

Table~\ref{tab:art2mus_train_results} reports objective metrics computed during training for different visual encoders and token counts. With ImageBind, IBSc consistently improved with more tokens, peaking at 16 tokens per image. CLIP did not exhibit a comparable monotonic trend. FAD and KL-Div remained stable across configurations. These results indicate that ImageBind’s higher-dimensional and multimodal embedding space benefits from increased token capacity, enabling richer conditioning signals.

\begin{table*}[b]
    \centering
    \caption{Average training metrics for different visual encoders and token counts. Arrows indicate whether higher ($\uparrow$) or lower ($\downarrow$) values are desirable. Best results per encoder are in \textbf{bold}; second-best are \underline{underlined}. Abbreviations: Tok/Img – Tokens per Image; Artw – Artwork; GeMus – Generated Music; GtMus – Ground-Truth Music.}
    \label{tab:art2mus_train_results}
    \begin{tabular}{c|cHH|Hcccc}
        \toprule
        \textbf{Image Encoder} &
        \textbf{Tok/Img} &
        \textbf{Artw} &
        \textbf{Artw Caption} &
        \textbf{Best Epoch} &
        \textbf{FAD $\downarrow$} &
        \textbf{KL-Div $\downarrow$} &
        \textbf{$\text{IBSc}_{\text{Artw--GeMus}} \uparrow$} &
        \textbf{$\text{IBSc}_{\text{GtMus--GeMus}} \uparrow$} \\
        \midrule

        \multirow{5}{*}{ImageBind} 
            & 4 & \multirow{3}{*}{\ding{51}} & \multirow{3}{*}{\ding{55}} & 14 & \textbf{19} & \textbf{0.00266} & 0.05716 & 0.19363 \\
            & 6 & & & 34 & \textbf{19} & \underline{0.00273} & 0.05693 & \underline{0.21398} \\
            & 8 & & & 15 & \textbf{19} & 0.00284 & \underline{0.05946} & 0.21341 \\
            & 10 & & & 17 & \textbf{19} & 0.00386  & 0.05288 & 0.18045 \\
            & 16 & & & 34 & \textbf{19} & 0.00279 & \textbf{0.05986} & \textbf{0.21661} \\
            
        \cmidrule(lr){1-9}

        \multirow{3}{*}{CLIP}
            & 4 & \multirow{5}{*}{\ding{51}} & \multirow{3}{*}{\ding{55}} & 15 & \underline{22} & 0.00320 & 0.05185 & \underline{0.18021} \\
            & 6 & & & 12 & 23 & 0.00317 & \textbf{0.05663} & 0.16921 \\
            & 8 & & & 12 & \underline{22} & 0.00310 & 0.05130 & 0.17757 \\
            & 10 & & & 34 & \textbf{21} & \textbf{0.00307} & 0.05378 & \textbf{0.19169} \\
            & 16 & & & 12 & \underline{22} & \underline{0.00309} & \underline{0.05546} & 0.17374 \\
        \bottomrule
    \end{tabular}
\end{table*}

Afterwards, we compared the best $\ArtToMus$ configurations (ImageBind with 16 tokens and CLIP with 6 tokens) against AudioLDM~2~\citep{liu2024audioldm}, AudioLDM~2-Music~\citep{liu2024audioldm}, and Mozart’s Touch~\citep{li2025mozart}. AudioLDM~2 models use text as input, with AudioLDM~2 generating music, speech, or sound effects, and AudioLDM~2-Music focused on music. Mozart's Touch takes an image and produces a caption, which is then converted into text that ultimately guides music generation. In contrast, $\ArtToMus$ does not require text and does not convert the image into a description; it directly uses image information to guide music generation. For text-based baselines, $\ArtSound$ artwork captions were provided as input; for Mozart’s Touch, the same captions were first transformed into short musical descriptions using Qwen3-14B, following Mozart’s Touch default prompt, to match the model’s expected input format.

\begin{table*}[t]
    \centering
    \caption{Objective evaluation results for $\ArtToMus$ and baselines: AudioLDM~2 (sound effects, human speech, and music), AudioLDM~2-Music (music only), and Mozart's Touch (music only). Arrows indicate whether higher ($\uparrow$) or lower ($\downarrow$) values 
    are preferable. Best values per row block are in \textbf{bold}; 
    second-best are \underline{underlined}. 
    Abbreviations: Emb.\ Model – Image Embedding Model; 
    Tok/Img – Tokens per Image; 
    Artw – Artwork; GeMus – Generated Music; 
    GtMus – Ground-Truth Music.}
    \label{tab:avg_objective_results}
    \resizebox{1.\linewidth}{!}{
    \begin{tabular}{lcccccccc}
        \toprule
        \textbf{Model} &
        \textbf{Emb.\ Model} &
        \textbf{Train Size (hrs)} &
        \textbf{Tok/Img} &
        \textbf{FAD $\downarrow$} &
        \textbf{KL-Div $\downarrow$} &
        \textbf{$\text{IBSc}_{\text{Artw--GeMus}} \uparrow$} &
        \textbf{$\text{IBSc}_{\text{GtMus--GeMus}} \uparrow$}
        \\
        \midrule
        $\ArtToMus$ & ImageBind & \multirow{2}{*}{$\approx$2.6} & 16  
            & \textbf{13} & \underline{0.00296} & 0.04671 & \underline{0.20368} \\
        $\ArtToMus$ & CLIP &  & 6 & 15 & 0.00374 & 0.04173 & 0.16716 \\
            
        AudioLDM~2 & – & $\approx$1,150,000 & –  
            & \underline{14} & 0.00425 & \underline{0.09822} & 0.17412 \\
        AudioLDM~2-Music & – & $\approx$665,000 & –  
            & 15 & 0.08449 & \textbf{0.18683} & 0.18683 \\
        Mozart's Touch & – & $\approx$20,000 & –  
            & 19 & \textbf{0.00238} & 0.08348 & \textbf{0.21629} \\
        \bottomrule
    \end{tabular}
    }
\end{table*}

ImageBind-based $\ArtToMus$ achieved the lowest FAD (13), a competitive KL-Div (0.00296), and the second highest IBSc with ground-truth music (0.20368). In comparison, the CLIP-based variant shows slightly worse performance across all metrics, confirming that ImageBind embeddings better capture the cross-modal relationships between artwork and music. AudioLDM~2, trained on a massive dataset ($\approx$1,150,000h), achieved a low FAD (14) and competitive IBSc with artwork (0.09822), but its KL-Div is higher than $\ArtToMus$. AudioLDM~2-Music, despite being trained on fewer hours ($\approx$665,000h), achieved the highest Artw–GeMus IBSc (0.18683), indicating strong alignment with the artwork, but its KL-Div is substantially worse (0.08449). Mozart's Touch, trained on a smaller dataset ($\approx$20,000h), shows the lowest KL-Div (0.00238) and the highest GtMus–GeMus IBSc (0.21629), highlighting its effectiveness at capturing characteristics of ground-truth music. However, it has the highest FAD (19).

It is important to note that text-based baselines benefit from explicit semantic conditioning derived from captions and language models. In this sense, they should be interpreted as upper bounds for alignment under semantic supervision rather than as direct competitors to a purely visual-conditioned system. Overall, $\ArtToMus$ achieves a strong trade-off between perceptual quality and cross-modal coherence under a strictly harder setting, since it performs direct artwork-to-music generation without text-derived semantic supervision.

A subjective evaluation of music generated by $\ArtToMus$ was conducted to analyze its qualitative characteristics, involving fifteen participants (3 musicians, 2 visual artists, and 10 without an artistic background). Each participant evaluated 15 artworks by answering the four questions in Table~\ref{tab:subj_eval_questions}, scoring audio quality, expressiveness, pertinence, and alignment, and selecting a preferred track for each artwork. Results are summarized in Table~\ref{tab:avg_subjective_results}.

\begin{table}[b]
    \centering
    \caption{Questions used for the subjective evaluation. Participants scored each generated track on audio quality, expressiveness, pertinence, and alignment.}
    \label{tab:subj_eval_questions}
    \begin{tabular}{c|l|c}
        \toprule
        \textbf{ID} & \textbf{Question} & \textbf{Scale} \\
        \midrule
        Q1 & How well does the music sound? & 1--5 \\
        Q2 & How expressive or emotionally engaging is the music? & 1--5 \\
        Q3 & How relevant is the music to the artwork? & 1--5 \\
        Q4 & How well does the music reflect the mood/content of the artwork? & 1--5 \\
        \bottomrule
    \end{tabular}
\end{table}

\begin{table*}[t]
    \centering
    \caption{Subjective evaluation results (mean scores across 15 participants). Each track was evaluated on audio quality, expressiveness, pertinence to the artwork, and alignment with the artwork. For each model, we computed each participant's average rating and averaged these across all participants to obtain the final score for each question. Best values are in \textbf{bold}.} 
    \label{tab:avg_subjective_results}
    \resizebox{1.\linewidth}{!}{
    \begin{tabular}{lcccc}
        \toprule
        \textbf{Model}
        & \textbf{Audio Quality (Q1) $\uparrow$}
        & \textbf{Expressiveness (Q2) $\uparrow$}
        & \textbf{Pertinence (Q3) $\uparrow$}
        & \textbf{Alignment (Q4) $\uparrow$} 
        \\
        \midrule
        $\ArtToMus$ (CLIP only artws) & 2.72 & \textbf{3.10} & \textbf{2.91} & \textbf{2.88} \\
        $\ArtToMus$ (ImageBind only artws) & \textbf{2.74} & 3.02 & 2.87 & 2.79 \\
        \bottomrule
    \end{tabular}
    }
\end{table*}

Both $\ArtToMus$ configurations show a consistent trade-off between audio quality and semantic alignment. ImageBind-based generations slightly outperform in audio quality, suggesting that their timbral characteristics are clearer or more stable. At the same time, CLIP-based outputs score higher in expressiveness, pertinence, and alignment, indicating a better capture of the artwork’s semantic and emotional content. Overall, $\ArtToMus$ demonstrates competitive performance in aligning generated music with visual artworks while maintaining reasonable audio quality.

\subsection{Discussion}

$\ArtToMus$ performs direct artwork-to-music generation without relying on text conditioning, which defines a strictly harder setting than caption-based baselines. Therefore, lower absolute scores compared to text-conditioned systems do not indicate inferior modeling, but rather reflect the increased difficulty of learning visual--audio correspondences without explicit semantic supervision. Despite this, $\ArtToMus$ achieves competitive perceptual quality and cross-modal alignment, and this result is notable given the absence of language-derived semantic cues.

Importantly, text-conditioned pipelines inherently compress visual information into linguistic abstractions. Two artworks with similar captions but radically different visual structures may therefore collapse to similar musical outputs in text-based systems, since fine-grained compositional and stylistic cues that are difficult to verbalize are discarded during captioning. By operating directly in the visual embedding space, $\ArtToMus$ preserves structural information that would otherwise be lost through textual mediation.


Comprehensively, $\ArtToMus$ demonstrates that direct artwork-to-music generation is feasible at scale and establishes a foundation for future multimodal creative systems.

\section{Conclusion}
\label{sec:conclusion}

This work introduced $\ArtToMus$, a framework for direct artwork-to-music generation that removes the reliance on textual intermediates and instead learns visual–audio correspondences directly from image embeddings. Inspired by the generative paradigm of latent diffusion models, $\ArtToMus$ reformulates the conditioning interface through a \textit{Visual Conditioning Extractor} and an \textit{Image Aligner}, enabling pretrained audio diffusion backbones to operate under purely visual conditioning. Alongside the model, we presented $\ArtSound$, a large-scale dataset of 105k artwork–music pairs enriched with image and audio captions, together with two alignment metrics, $\ICscore$ and $\ACscore$, to assess caption reliability. Beyond providing training data, $\ArtSound$ establishes a benchmark for studying cross-modal creative generation at scale.

Experimental results demonstrate that direct visual-to-music generation is feasible and can achieve competitive perceptual quality and meaningful cross-modal alignment. While text-conditioned systems benefit from explicit semantic supervision and therefore represent upper bounds under language-mediated conditioning, $\ArtToMus$ addresses a strictly harder problem by learning correspondences directly from visual structure. This distinction is fundamental: compressing artworks into captions inevitably discards fine-grained compositional and stylistic information, whereas direct visual conditioning preserves these cues within the generative process.

More broadly, this work establishes direct artwork-to-music generation as a viable research direction in multimodal generative AI. Future work includes scaling the dataset to improve stylistic coverage, exploring richer visual tokenization and adaptive LoA representations, integrating retrieval-augmented or hierarchical conditioning mechanisms, and incorporating explainable AI techniques~\citep{castellano2024using} to better understand how visual features influence musical structure. Enabling user-guided editing through visual or hybrid visual–textual controls also represents a promising direction for interactive and personalized creative systems.

\begin{acks}

We acknowledge ISCRA for granting access to the LEONARDO supercomputer, owned by the EuroHPC Joint Undertaking and hosted by CINECA (Italy).

The research of Ivan Rinaldi is supported by a Ph.D.\ fellowship within the project ``Patti Territoriali dell'Alta Formazione per le Imprese'', University of Bari Aldo Moro, Ph.D.\ program ``Advanced Generative Models for Conversion between Images, Text, and Audio'', co-funded by AI Factory S.r.l.\ (CUP F61B23000370006).

The research of Nicola Fanelli is supported by a Ph.D.\ fellowship under the Italian ``D.M. n. 118/23'', within the National Recovery and Resilience Plan, Mission 4, Component 2, Investment 4.1, Ph.D.\ project ``Analysis and valorization of digitized artistic heritage using Artificial Intelligence techniques'' (CUP H91I23000690007).

\end{acks}

\bibliographystyle{ACM-Reference-Format}
\bibliography{main}

@inproceedings{rinaldi2024art2mus,
	title        = {{Art2Mus: Bridging Visual Arts and Music through Cross-Modal Generation}},
	author       = {Rinaldi, Ivan and Fanelli, Nicola and Castellano, Giovanna and Vessio, Gennaro},
	year         = 2024,
	booktitle    = {Computer Vision - {ECCV} 2024 Workshops - Milan, Italy, September 29-October 4, 2024, Proceedings, Part {V}},
	publisher    = {Springer},
	series       = {Lecture Notes in Computer Science},
	volume       = 15627,
	pages        = {173--186}
}

@inproceedings{song2025musflow,
	title        = {{MusFlow: Multimodal Music Generation via Conditional Flow Matching}},
	author       = {Song, Jiahao and Wang, Yu-Zhao},
	year         = 2025,
	booktitle    = {Proceedings of the 33rd {ACM} International Conference on Multimedia, {MM} 2025, Dublin, Ireland, October 27-31, 2025},
	publisher    = {{ACM}},
	pages        = {10200--10209}
}

@article{agostinelli2023musiclm,
	title        = {{Musiclm: Generating music from text}},
	author       = {Agostinelli, Andrea and Denk, Timo I and Borsos, Zal{\'a}n and Engel, Jesse and Verzetti, Mauro and Caillon, Antoine and Huang, Qingqing and Jansen, Aren and Roberts, Adam and Tagliasacchi, Marco and others},
	year         = 2023,
	journal      = {arXiv preprint arXiv:2301.11325},
	volume       = {abs/2301.11325}
}

@inproceedings{copet2023simple,
	title        = {{Simple and controllable music generation}},
	author       = {Copet, Jade and Kreuk, Felix and Gat, Itai and Remez, Tal and Kant, David and Synnaeve, Gabriel and Adi, Yossi and D{\'e}fossez, Alexandre},
	year         = 2023,
	booktitle    = {Advances in Neural Information Processing Systems 36: Annual Conference on Neural Information Processing Systems 2023, NeurIPS 2023, New Orleans, LA, USA, December 10 - 16, 2023},
	volume       = 36,
	pages        = {47704--47720}
}

@article{liu2024audioldm,
	title        = {{AudioLDM 2: Learning Holistic Audio Generation With Self-Supervised Pretraining}},
	author       = {Liu, Haohe and Yuan, Yi and Liu, Xubo and Mei, Xinhao and Kong, Qiuqiang and Tian, Qiao and Wang, Yuping and Wang, Wenwu and Wang, Yuxuan and Plumbley, Mark D.},
	year         = 2024,
	journal      = {IEEE/ACM Transactions on Audio, Speech, and Language Processing},
	publisher    = {IEEE},
	volume       = 32,
	pages        = {2871--2883}
}

@article{castellano2022leveraging,
	title        = {{Leveraging knowledge graphs and deep learning for automatic art analysis}},
	author       = {Castellano, Giovanna and Digeno, Vincenzo and Sansaro, Giovanni and Vessio, Gennaro},
	year         = 2022,
	journal      = {Knowledge-Based Systems},
	publisher    = {Elsevier},
	volume       = 248,
	pages        = 108859
}

@inproceedings{defferrard2016fma,
	title        = {{FMA: A dataset for music analysis}},
	author       = {Defferrard, Micha{\"e}l and Benzi, Kirell and Vandergheynst, Pierre and Bresson, Xavier},
	year         = 2017,
	booktitle    = {Proceedings of the 18th International Society for Music Information Retrieval Conference, {ISMIR} 2017, Suzhou, China, October 23-27, 2017},
	pages        = {316--323}
}

@inproceedings{sarto2023positive,
	title        = {{Positive-augmented contrastive learning for image and video captioning evaluation}},
	author       = {Sarto, Sara and Barraco, Manuele and Cornia, Marcella and Baraldi, Lorenzo and Cucchiara, Rita},
	year         = 2023,
	booktitle    = {{IEEE/CVF} Conference on Computer Vision and Pattern Recognition, {CVPR} 2023, Vancouver, BC, Canada, June 17-24, 2023},
	publisher    = {{IEEE}},
	pages        = {6914--6924}
}

@inproceedings{hessel2021clipscore,
	title        = {{Clipscore: A reference-free evaluation metric for image captioning}},
	author       = {Hessel, Jack and Holtzman, Ari and Forbes, Maxwell and Le Bras, Ronan and Choi, Yejin},
	year         = 2021,
	booktitle    = {Proceedings of the 2021 Conference on Empirical Methods in Natural Language Processing, {EMNLP} 2021, Virtual Event / Punta Cana, Dominican Republic, 7-11 November, 2021},
	publisher    = {Association for Computational Linguistics},
	pages        = {7514--7528}
}

@inproceedings{liu2023visual,
	title        = {{Visual instruction tuning}},
	author       = {Liu, Haotian and Li, Chunyuan and Wu, Qingyang and Lee, Yong Jae},
	year         = 2023,
	booktitle    = {Advances in Neural Information Processing Systems 36: Annual Conference on Neural Information Processing Systems 2023, NeurIPS 2023, New Orleans, LA, USA, December 10 - 16, 2023},
	volume       = 36,
	pages        = {34892--34916}
}

@inproceedings{doh2023lp,
	title        = {{Lp-musiccaps: Llm-based pseudo music captioning}},
	author       = {Doh, SeungHeon and Choi, Keunwoo and Lee, Jongpil and Nam, Juhan},
	year         = 2023,
	booktitle    = {Proceedings of the 24th International Society for Music Information Retrieval Conference, {ISMIR} 2023, Milan, Italy, November 5-9, 2023},
	pages        = {409--416}
}

@inproceedings{lin2004rouge,
	title        = {{Rouge: A package for automatic evaluation of summaries}},
	author       = {Lin, Chin-Yew},
	year         = 2004,
	booktitle    = {Text summarization branches out},
	publisher    = {Association for Computational Linguistics},
	address      = {Barcelona, Spain},
	pages        = {74--81}
}

@inproceedings{zhang2020bertscore,
	title        = {{Bertscore: Evaluating text generation with bert}},
	author       = {Zhang, Tianyi and Kishore, Varsha and Wu, Felix and Weinberger, Kilian Q and Artzi, Yoav},
	year         = 2020,
	booktitle    = {8th International Conference on Learning Representations, {ICLR} 2020, Addis Ababa, Ethiopia, April 26-30, 2020},
	publisher    = {OpenReview.net}
}

@article{liu2019roberta,
	title        = {{Roberta: A robustly optimized bert pretraining approach}},
	author       = {Liu, Yinhan and Ott, Myle and Goyal, Naman and Du, Jingfei and Joshi, Mandar and Chen, Danqi and Levy, Omer and Lewis, Mike and Zettlemoyer, Luke and Stoyanov, Veselin},
	year         = 2019,
	journal      = {arXiv preprint arXiv:1907.11692}
}

@inproceedings{warner2025smarter,
	title        = {{Smarter, better, faster, longer: A modern bidirectional encoder for fast, memory efficient, and long context finetuning and inference}},
	author       = {Warner, Benjamin and Chaffin, Antoine and Clavi{\'e}, Benjamin and Weller, Orion and Hallstr{\"o}m, Oskar and Taghadouini, Said and Gallagher, Alexis and Biswas, Raja and Ladhak, Faisal and Aarsen, Tom and others},
	year         = 2025,
	booktitle    = {Proceedings of the 63rd Annual Meeting of the Association for Computational Linguistics (Volume 1: Long Papers), {ACL} 2025, Vienna, Austria, July 27 - August 1, 2025},
	publisher    = {Association for Computational Linguistics},
	pages        = {2526--2547}
}

@inproceedings{kilgour2018fr,
	title        = {{Fr$\backslash$'echet audio distance: A metric for evaluating music enhancement algorithms}},
	author       = {Kilgour, Kevin and Zuluaga, Mauricio and Roblek, Dominik and Sharifi, Matthew},
	year         = 2019,
	booktitle    = {20th Annual Conference of the International Speech Communication Association, Interspeech 2019, Graz, Austria, September 15-19, 2019},
	publisher    = {{ISCA}},
	pages        = {2350--2354}
}

@inproceedings{girdhar2023imagebind,
	title        = {{Imagebind: One embedding space to bind them all}},
	author       = {Girdhar, Rohit and El-Nouby, Alaaeldin and Liu, Zhuang and Singh, Mannat and Alwala, Kalyan Vasudev and Joulin, Armand and Misra, Ishan},
	year         = 2023,
	booktitle    = {Proceedings of the IEEE/CVF conference on computer vision and pattern recognition},
	publisher    = {IEEE},
	pages        = {15180--15190}
}

@inproceedings{rombach2022high,
	title        = {{High-Resolution Image Synthesis with Latent Diffusion Models}},
	author       = {Rombach, Robin and Blattmann, Andreas and Lorenz, Dominik and Esser, Patrick and Ommer, Bj{\"o}rn},
	year         = 2022,
	booktitle    = {{IEEE/CVF} Conference on Computer Vision and Pattern Recognition, {CVPR} 2022, New Orleans, LA, USA, June 18-24, 2022},
	publisher    = {{IEEE}},
	pages        = {10674--10685}
}

@inproceedings{li2023lavis,
	title        = {{LAVIS: A one-stop library for language-vision intelligence}},
	author       = {Li, Dongxu and Li, Junnan and Le, Hung and Wang, Guangsen and Savarese, Silvio and Hoi, Steven CH},
	year         = 2023,
	booktitle    = {Proceedings of the 61st Annual Meeting of the Association for Computational Linguistics: System Demonstrations, {ACL} 2023, Toronto, Canada, July 10-12, 2023},
	publisher    = {Association for Computational Linguistics},
	pages        = {31--41}
}

@inproceedings{castellano2023exploring,
	title        = {{Exploring the Synergy Between Vision-Language Pretraining and ChatGPT for Artwork Captioning: A Preliminary Study}},
	author       = {Castellano, Giovanna and Fanelli, Nicola and Scaringi, Raffaele and Vessio, Gennaro},
	year         = 2023,
	booktitle    = {International Conference on Image Analysis and Processing},
	publisher    = {Springer},
	pages        = {309--321}
}

@inproceedings{bai2021explain,
	title        = {{Explain me the painting: Multi-topic knowledgeable art description generation}},
	author       = {Bai, Zechen and Nakashima, Yuta and Garcia, Noa},
	year         = 2021,
	booktitle    = {Proceedings of the IEEE/CVF international conference on computer vision},
	publisher    = {{IEEE}},
	pages        = {5422--5432}
}

@article{liu2026mumu,
	title        = {{Mumu-llama: Multi-modal music understanding and generation via large language models}},
	author       = {Liu, Shansong and Hussain, Atin Sakkeer and Wu, Qilong and Sun, Chenshuo and Shan, Ying},
	year         = 2026,
	journal      = {Expert Syst. Appl.},
	volume       = 305,
	pages        = 130688
}

@inproceedings{deng2024musilingo,
	title        = {{Musilingo: Bridging music and text with pre-trained language models for music captioning and query response}},
	author       = {Deng, Zihao and Ma, Yinghao and Liu, Yudong and Guo, Rongchen and Zhang, Ge and Chen, Wenhu and Huang, Wenhao and Benetos, Emmanouil},
	year         = 2024,
	booktitle    = {Findings of the Association for Computational Linguistics: NAACL 2024},
	publisher    = {Association for Computational Linguistics},
	pages        = {3643--3655}
}

@inproceedings{hang2023efficient,
	title        = {{Efficient diffusion training via min-snr weighting strategy}},
	author       = {Hang, Tiankai and Gu, Shuyang and Li, Chen and Bao, Jianmin and Chen, Dong and Hu, Han and Geng, Xin and Guo, Baining},
	year         = 2023,
	booktitle    = {{IEEE/CVF} International Conference on Computer Vision, {ICCV} 2023, Paris, France, October 1-6, 2023},
	publisher    = {{IEEE}},
	pages        = {7407--7417}
}

@article{chung2024scaling,
	title        = {{Scaling instruction-finetuned language models}},
	author       = {Chung, Hyung Won and Hou, Le and Longpre, Shayne and Zoph, Barret and Tay, Yi and Fedus, William and Li, Yunxuan and Wang, Xuezhi and Dehghani, Mostafa and Brahma, Siddhartha and others},
	year         = 2024,
	journal      = {Journal of Machine Learning Research},
	volume       = 25,
	pages        = {70:1--70:53}
}

@article{huang2022masked,
	title        = {{Masked autoencoders that listen}},
	author       = {Huang, Po-Yao and Xu, Hu and Li, Juncheng and Baevski, Alexei and Auli, Michael and Galuba, Wojciech and Metze, Florian and Feichtenhofer, Christoph},
	year         = 2022,
	journal      = {Advances in Neural Information Processing Systems},
	volume       = 35,
	pages        = {28708--28720}
}

@article{radford2019language,
	title        = {{Language models are unsupervised multitask learners}},
	author       = {Radford, Alec and Wu, Jeffrey and Child, Rewon and Luan, David and Amodei, Dario and Sutskever, Ilya and others},
	year         = 2019,
	journal      = {OpenAI blog},
	volume       = 1,
	number       = 8,
	pages        = 9
}

@inproceedings{radford2021learning,
	title        = {{Learning transferable visual models from natural language supervision}},
	author       = {Radford, Alec and Kim, Jong Wook and Hallacy, Chris and Ramesh, Aditya and Goh, Gabriel and Agarwal, Sandhini and Sastry, Girish and Askell, Amanda and Mishkin, Pamela and Clark, Jack and others},
	year         = 2021,
	booktitle    = {Proceedings of the 38th International Conference on Machine Learning, {ICML} 2021, 18-24 July 2021, Virtual Event},
	publisher    = {{PMLR}},
	series       = {Proceedings of Machine Learning Research},
	volume       = 139,
	pages        = {8748--8763}
}

@inproceedings{li2025mozart,
	title        = {{Mozart’s Touch: a lightweight multimodal music generation framework based on pre-trained large models}},
	author       = {Li, Jiajun and Xu, Tianze and Chen, Xuesong and Yao, Xinrui and Han, Jingchou and Liu, Shuchang},
	year         = 2025,
	booktitle    = {International Conference on AI-Generated Content (AIGC 2024)},
	publisher    = {SPIE},
	volume       = 13649,
	pages        = {198--207}
}

@inproceedings{deshmukh2023pengi,
	title        = {{Pengi: An audio language model for audio tasks}},
	author       = {Deshmukh, Soham and Elizalde, Benjamin and Singh, Rita and Wang, Huaming},
	year         = 2023,
	booktitle    = {Advances in Neural Information Processing Systems 36: Annual Conference on Neural Information Processing Systems 2023, NeurIPS 2023, New Orleans, LA, USA, December 10 - 16, 2023},
	volume       = 36,
	pages        = {18090--18108}
}

@inproceedings{mei2021audio,
	title        = {{Audio Captioning Transformer}},
	author       = {Mei, Xinhao and Liu, Xubo and Huang, Qiushi and Plumbley, Mark D. and Wang, Wenwu},
	year         = 2021,
	month        = {November},
	booktitle    = {Proceedings of the 6th Detection and Classification of Acoustic Scenes and Events 2021 Workshop (DCASE2021)},
	address      = {Barcelona, Spain},
	pages        = {211--215}
}

@inproceedings{mitchell2012midge,
	title        = {{Midge: Generating image descriptions from computer vision detections}},
	author       = {Mitchell, Margaret and Dodge, Jesse and Goyal, Amit and Yamaguchi, Kota and Stratos, Karl and Han, Xufeng and Mensch, Alyssa and Berg, Alexander and Berg, Tamara and Daum{\'e} III, Hal},
	year         = 2012,
	booktitle    = {{EACL} 2012, 13th Conference of the European Chapter of the Association for Computational Linguistics, Avignon, France, April 23-27, 2012},
	publisher    = {The Association for Computer Linguistics},
	pages        = {747--756}
}

@inproceedings{manco2021muscaps,
	title        = {{Muscaps: Generating captions for music audio}},
	author       = {Manco, Ilaria and Benetos, Emmanouil and Quinton, Elio and Fazekas, Gy{\"o}rgy},
	year         = 2021,
	booktitle    = {2021 International Joint Conference on Neural Networks (IJCNN)},
	publisher    = {{IEEE}},
	pages        = {1--8}
}

@inproceedings{melechovsky2024mustango,
	title        = {{Mustango: Toward controllable text-to-music generation}},
	author       = {Melechovsky, Jan and Guo, Zixun and Ghosal, Deepanway and Majumder, Navonil and Herremans, Dorien and Poria, Soujanya},
	year         = 2024,
	booktitle    = {Proceedings of the 2024 Conference of the North American Chapter of the Association for Computational Linguistics: Human Language Technologies (Volume 1: Long Papers), {NAACL} 2024, Mexico City, Mexico, June 16-21, 2024},
	publisher    = {Association for Computational Linguistics},
	pages        = {8293--8316}
}

@inproceedings{liu2025musiceval,
	title        = {{Musiceval: A generative music dataset with expert ratings for automatic text-to-music evaluation}},
	author       = {Liu, Cheng and Wang, Hui and Zhao, Jinghua and Zhao, Shiwan and Bu, Hui and Xu, Xin and Zhou, Jiaming and Sun, Haoqin and Qin, Yong},
	year         = 2025,
	booktitle    = {2025 {IEEE} International Conference on Acoustics, Speech and Signal Processing, {ICASSP} 2025, Hyderabad, India, April 6-11, 2025},
	publisher    = {{IEEE}},
	pages        = {1--5}
}

@article{lu2024artcap,
	title        = {{Artcap: A dataset for image captioning of fine art paintings}},
	author       = {Lu, Yue and Guo, Chao and Dai, Xingyuan and Wang, Fei-Yue},
	year         = 2024,
	journal      = {IEEE Transactions on Computational Social Systems},
	publisher    = {IEEE},
	volume       = 11,
	number       = 1,
	pages        = {576--587}
}

@inproceedings{cetinic2021iconographic,
	title        = {{Iconographic image captioning for artworks}},
	author       = {Cetinic, Eva},
	year         = 2021,
	booktitle    = {International Conference on Pattern Recognition},
	publisher    = {Springer},
	pages        = {502--516}
}

@inproceedings{achlioptas2021artemis,
	title        = {{Artemis: Affective language for visual art}},
	author       = {Achlioptas, Panos and Ovsjanikov, Maks and Haydarov, Kilichbek and Elhoseiny, Mohamed and Guibas, Leonidas J},
	year         = 2021,
	booktitle    = {Proceedings of the IEEE/CVF Conference on Computer Vision and Pattern Recognition},
	publisher    = {Computer Vision Foundation / {IEEE}},
	pages        = {11569--11579}
}

@article{sharma2024m2m,
	title        = {{M2m-gen: A multimodal framework for automated background music generation in japanese manga using large language models}},
	author       = {Sharma, Megha and Haseeb, Muhammad Taimoor and Xia, Gus and Tsuruoka, Yoshimasa},
	year         = 2024,
	journal      = {arXiv preprint arXiv:2410.09928}
}

@article{liu2023m,
	title        = {{M\textsuperscript{2}M-Gen: Multi-modal Music Understanding and Generation with the Power of Large Language Models}},
	author       = {Liu, Shansong and Hussain, Atin Sakkeer and Sun, Chenshuo and Shan, Ying},
	year         = 2023,
	journal      = {arXiv preprint arXiv:2311.11255}
}

@inproceedings{huang2023make,
	title        = {{Make-an-audio: Text-to-audio generation with prompt-enhanced diffusion models}},
	author       = {Huang, Rongjie and Huang, Jiawei and Yang, Dongchao and Ren, Yi and Liu, Luping and Li, Mingze and Ye, Zhenhui and Liu, Jinglin and Yin, Xiang and Zhao, Zhou},
	year         = 2023,
	booktitle    = {International Conference on Machine Learning, {ICML} 2023, 23-29 July 2023, Honolulu, Hawaii, {USA}},
	publisher    = {{PMLR}},
	series       = {Proceedings of Machine Learning Research},
	volume       = 202,
	pages        = {13916--13932}
}

@inproceedings{ge2023expressive,
	title        = {{Expressive text-to-image generation with rich text}},
	author       = {Ge, Songwei and Park, Taesung and Zhu, Jun-Yan and Huang, Jia-Bin},
	year         = 2023,
	booktitle    = {Proceedings of the IEEE/CVF International Conference on Computer Vision},
	publisher    = {{IEEE}},
	pages        = {7545--7556}
}

@inproceedings{zhu2023conditional,
	title        = {{Conditional text image generation with diffusion models}},
	author       = {Zhu, Yuanzhi and Li, Zhaohai and Wang, Tianwei and He, Mengchao and Yao, Cong},
	year         = 2023,
	booktitle    = {Proceedings of the IEEE/CVF Conference on Computer Vision and Pattern Recognition},
	publisher    = {{IEEE}},
	pages        = {14235--14245}
}

@inproceedings{chowdhury2024melfusion,
	title        = {{Melfusion: Synthesizing music from image and language cues using diffusion models}},
	author       = {Chowdhury, Sanjoy and Nag, Sayan and Joseph, KJ and Srinivasan, Balaji Vasan and Manocha, Dinesh},
	year         = 2024,
	booktitle    = {Proceedings of the IEEE/CVF Conference on Computer Vision and Pattern Recognition},
	publisher    = {{IEEE}},
	pages        = {26826--26835}
}

@inproceedings{sheffer2023hear,
	title        = {{I hear your true colors: Image guided audio generation}},
	author       = {Sheffer, Roy and Adi, Yossi},
	year         = 2023,
	booktitle    = {ICASSP 2023-2023 IEEE International Conference on Acoustics, Speech and Signal Processing (ICASSP)},
	publisher    = {IEEE},
	pages        = {1--5}
}

@article{borsos2023audiolm,
	title        = {{Audiolm: a language modeling approach to audio generation}},
	author       = {Borsos, Zal{\'a}n and Marinier, Rapha{\"e}l and Vincent, Damien and Kharitonov, Eugene and Pietquin, Olivier and Sharifi, Matt and Roblek, Dominik and Teboul, Olivier and Grangier, David and Tagliasacchi, Marco and others},
	year         = 2023,
	journal      = {IEEE/ACM transactions on audio, speech, and language processing},
	publisher    = {IEEE Press},
	volume       = 31,
	pages        = {2523--2533}
}

@article{liaoimage,
	title        = {{An Image-to-Music Generation Framework Powered by An Algorithm-Driven Music Core}},
	author       = {Liao, Callie C and Liao, Duoduo and Zhang, Ellie L},
	year         = 2025
}

@inproceedings{castellano2024using,
	title        = {{Using LLMs to explain AI-generated art classification via Grad-CAM heatmaps}},
	author       = {Castellano, Giovanna and Miccoli, MARIA GRAZIA and Scaringi, Raffaele and Vessio, Gennaro and Zaza, Gianluca and others},
	year         = 2024,
	booktitle    = {Proceedings of 5th Italian Workshop on Explainable Artificial Intelligence, co-located with the 23rd International Conference of the Italian Association for Artificial Intelligence, Bolzano, Italy},
	publisher    = {CEUR-WS.org},
	pages        = {65--74}
}

@inproceedings{gemmeke2017audio,
	title        = {{Audio set: An ontology and human-labeled dataset for audio events}},
	author       = {Gemmeke, Jort F and Ellis, Daniel PW and Freedman, Dylan and Jansen, Aren and Lawrence, Wade and Moore, R Channing and Plakal, Manoj and Ritter, Marvin},
	year         = 2017,
	booktitle    = {2017 IEEE international conference on acoustics, speech and signal processing (ICASSP)},
	publisher    = {IEEE},
	pages        = {776--780}
}

@inproceedings{plummer2015flickr30k,
	title        = {{Flickr30k entities: Collecting region-to-phrase correspondences for richer image-to-sentence models}},
	author       = {Plummer, Bryan A and Wang, Liwei and Cervantes, Chris M and Caicedo, Juan C and Hockenmaier, Julia and Lazebnik, Svetlana},
	year         = 2015,
	booktitle    = {2015 {IEEE} International Conference on Computer Vision, {ICCV} 2015, Santiago, Chile, December 7-13, 2015},
	publisher    = {{IEEE} Computer Society},
	pages        = {2641--2649}
}

@article{chen2015microsoft,
	title        = {{Microsoft coco captions: Data collection and evaluation server}},
	author       = {Chen, Xinlei and Fang, Hao and Lin, Tsung-Yi and Vedantam, Ramakrishna and Gupta, Saurabh and Doll{\'a}r, Piotr and Zitnick, C Lawrence},
	year         = 2015,
	journal      = {arXiv preprint arXiv:1504.00325},
	volume       = {abs/1504.00325}
}

@inproceedings{kim2019audiocaps,
	title        = {{Audiocaps: Generating captions for audios in the wild}},
	author       = {Kim, Chris Dongjoo and Kim, Byeongchang and Lee, Hyunmin and Kim, Gunhee},
	year         = 2019,
	booktitle    = {Proceedings of the 2019 Conference of the North American Chapter of the Association for Computational Linguistics: Human Language Technologies, {NAACL-HLT} 2019, Minneapolis, MN, USA, June 2-7, 2019, Volume 1 (Long and Short Papers)},
	publisher    = {Association for Computational Linguistics},
	pages        = {119--132}
}

@article{bai2025audiosetcaps,
	title        = {{Audiosetcaps: An enriched audio-caption dataset using automated generation pipeline with large audio and language models}},
	author       = {Bai, Jisheng and Liu, Haohe and Wang, Mou and Shi, Dongyuan and Wang, Wenwu and Plumbley, Mark D and Gan, Woon-Seng and Chen, Jianfeng},
	year         = 2025,
	journal      = {IEEE Transactions on Audio, Speech and Language Processing},
	publisher    = {IEEE}
}

@inproceedings{wang2023diffusiondb,
	title        = {{Diffusiondb: A large-scale prompt gallery dataset for text-to-image generative models}},
	author       = {Wang, Zijie J and Montoya, Evan and Munechika, David and Yang, Haoyang and Hoover, Benjamin and Chau, Duen Horng},
	year         = 2023,
	booktitle    = {Proceedings of the 61st Annual Meeting of the Association for Computational Linguistics (Volume 1: Long Papers), {ACL} 2023, Toronto, Canada, July 9-14, 2023},
	publisher    = {Association for Computational Linguistics},
	pages        = {893--911}
}

@inproceedings{srinivasan2021wit,
	title        = {{Wit: Wikipedia-based image text dataset for multimodal multilingual machine learning}},
	author       = {Srinivasan, Krishna and Raman, Karthik and Chen, Jiecao and Bendersky, Michael and Najork, Marc},
	year         = 2021,
	booktitle    = {{SIGIR} '21: The 44th International {ACM} {SIGIR} Conference on Research and Development in Information Retrieval, Virtual Event, Canada, July 11-15, 2021},
	publisher    = {{ACM}},
	pages        = {2443--2449}
}

@article{shih2022theme,
	title        = {{Theme transformer: Symbolic music generation with theme-conditioned transformer}},
	author       = {Shih, Yi-Jen and Wu, Shih-Lun and Zalkow, Frank and M{\"u}ller, Meinard and Yang, Yi-Hsuan},
	year         = 2022,
	journal      = {IEEE Transactions on Multimedia},
	publisher    = {IEEE},
	volume       = 25,
	pages        = {3495--3508}
}

@article{hu2023beauty,
	title        = {{The beauty of repetition: an algorithmic composition model with motif-level repetition generator and outline-to-music generator in symbolic music generation}},
	author       = {Hu, Zhejing and Ma, Xiao and Liu, Yan and Chen, Gong and Liu, Yongxu and Dannenberg, Roger B},
	year         = 2023,
	journal      = {IEEE Transactions on Multimedia},
	publisher    = {IEEE},
	volume       = 26,
	pages        = {4320--4333}
}

@article{ma2024symbolic,
	title        = {{Symbolic music generation from graph-learning-based preference modeling and textual queries}},
	author       = {Ma, Xichu and Wang, Yuchen and Wang, Ye},
	year         = 2024,
	journal      = {IEEE Transactions on Multimedia},
	publisher    = {IEEE},
	volume       = 26,
	pages        = {10545--10558}
}

@inproceedings{tan2020automated,
	title        = {{Automated Music Generation for Visual Art through Emotion.}},
	author       = {Tan, Xiaodong and Antony, Mathis and Kong, H},
	year         = 2020,
	booktitle    = {Proceedings of the Eleventh International Conference on Computational Creativity, {ICCC} 2020, Coimbra, Portugal, September 7-11, 2020},
	publisher    = {Association for Computational Creativity {(ACC)}},
	pages        = {247--250}
}

@inproceedings{kong2020hifi,
	title        = {{Hifi-gan: Generative adversarial networks for efficient and high fidelity speech synthesis}},
	author       = {Kong, Jungil and Kim, Jaehyeon and Bae, Jaekyoung},
	year         = 2020,
	booktitle    = {Advances in Neural Information Processing Systems 33: Annual Conference on Neural Information Processing Systems 2020, NeurIPS 2020, December 6-12, 2020, virtual},
	volume       = 33,
	pages        = {17022--17033}
}

@inproceedings{zhao2020emotion,
	title        = {{Emotion-based end-to-end matching between image and music in valence-arousal space}},
	author       = {Zhao, Sicheng and Li, Yaxian and Yao, Xingxu and Nie, Weizhi and Xu, Pengfei and Yang, Jufeng and Keutzer, Kurt},
	year         = 2020,
	booktitle    = {{MM} '20: The 28th {ACM} International Conference on Multimedia, Virtual Event / Seattle, WA, USA, October 12-16, 2020},
	publisher    = {{ACM}},
	pages        = {2945--2954}
}

@article{hisariya2024bridging,
	title        = {{Bridging Paintings and Music--Exploring Emotion based Music Generation through Paintings}},
	author       = {Hisariya, Tanisha and Zhang, Huan and Liang, Jinhua},
	year         = 2024,
	journal      = {arXiv preprint arXiv:2409.07827}
}

@misc{wang2025visiontomusicgenerationsurvey,
	title        = {{Vision-to-Music Generation: A Survey}},
	author       = {Zhaokai Wang and Chenxi Bao and Le Zhuo and Jingrui Han and Yang Yue and Yihong Tang and Victor Shea-Jay Huang and Yue Liao},
	year         = 2025,
	url          = {https://arxiv.org/abs/2503.21254},
	eprint       = {2503.21254},
	archiveprefix = {arXiv},
	primaryclass = {cs.CV}
}

@inproceedings{elizalde2023clap,
	title        = {{Clap learning audio concepts from natural language supervision}},
	author       = {Elizalde, Benjamin and Deshmukh, Soham and Al Ismail, Mahmoud and Wang, Huaming},
	year         = 2023,
	booktitle    = {ICASSP 2023-2023 IEEE International Conference on Acoustics, Speech and Signal Processing (ICASSP)},
	publisher    = {IEEE},
	pages        = {1--5}
}

@article{yang2025qwen3,
	title        = {{Qwen3 technical report}},
	author       = {Yang, An and Li, Anfeng and Yang, Baosong and Zhang, Beichen and Hui, Binyuan and Zheng, Bo and Yu, Bowen and Gao, Chang and Huang, Chengen and Lv, Chenxu and others},
	year         = 2025,
	journal      = {arXiv preprint arXiv:2505.09388},
	volume       = {abs/2505.09388}
}

@article{bai2025qwen2,
	title        = {{Qwen2.5-vl technical report}},
	author       = {Bai, Shuai and Chen, Keqin and Liu, Xuejing and Wang, Jialin and Ge, Wenbin and Song, Sibo and Dang, Kai and Wang, Peng and Wang, Shijie and Tang, Jun and others},
	year         = 2025,
	journal      = {arXiv preprint arXiv:2502.13923},
	volume       = {abs/2502.13923}
}

@article{xu2025qwen2,
	title        = {{Qwen2.5-omni technical report}},
	author       = {Xu, Jin and Guo, Zhifang and He, Jinzheng and Hu, Hangrui and He, Ting and Bai, Shuai and Chen, Keqin and Wang, Jialin and Fan, Yang and Dang, Kai and others},
	year         = 2025,
	journal      = {arXiv preprint arXiv:2503.20215},
	volume       = {abs/2503.20215}
}

@article{chu2024qwen2,
	title        = {{Qwen2-audio technical report}},
	author       = {Chu, Yunfei and Xu, Jin and Yang, Qian and Wei, Haojie and Wei, Xipin and Guo, Zhifang and Leng, Yichong and Lv, Yuanjun and He, Jinzheng and Lin, Junyang and others},
	year         = 2024,
	journal      = {arXiv preprint arXiv:2407.10759},
	volume       = {abs/2407.10759}
}

@inproceedings{
    song2021denoising,
    title={Denoising Diffusion Implicit Models},
    author={Jiaming Song and Chenlin Meng and Stefano Ermon},
    booktitle={International Conference on Learning Representations},
    year={2021},
    url={https://openreview.net/forum?id=St1giarCHLP}
}

@article{grattafiori2024llama,
  title={The llama 3 herd of models},
  author={Grattafiori, Aaron and Dubey, Abhimanyu and Jauhri, Abhinav and Pandey, Abhinav and Kadian, Abhishek and Al-Dahle, Ahmad and Letman, Aiesha and Mathur, Akhil and Schelten, Alan and Vaughan, Alex and others},
  journal={arXiv preprint arXiv:2407.21783},
  year         = {2024},
  volume       = {abs/2407.21783},
}

\appendix

\section{Captioning}
\label{abstud:captionig}

As described in the main text (Section~\ref{sec:dataset_creation}), we evaluated additional models for both image and audio captioning, but they were not included in the final pipeline (Figure~\ref{fig:artw_music_caption_pipelines}) due to issues related to caption length or content quality. The following subsections present sample captions generated by these discarded models for both tasks.

\subsection{Image Captioning}
\label{abstud:image_capt}

\begin{table*}[b]
\centering
\caption{Example of artwork captions generated using \textcolor{red!90!black}{\textbf{discarded models}}. LAVIS does not require a prompt. All remaining models were given the following instructions: ``You are an art expert. Provide a concise description of this artwork, addressing its subject matter, the mood or emotion it conveys, key artistic techniques and style, and its possible artistic period or school''.}
\small
\begin{tabularx}{\linewidth}{@{} >{\centering\arraybackslash}m{0.25\textwidth} | >{\arraybackslash}m{0.7\textwidth} @{}}
\toprule
\textbf{Image} & \textbf{Captions} \\
\midrule
    \begin{minipage}[c]{\linewidth}
        \centering
        \includegraphics[width=.9\linewidth]{figs/artworks/camille-pissarro_the-hills-of-thierceville.jpg} \\[0.5ex]
        The Hills of Thierceville \\ Camille Pissarro
        \vspace{0.25em}
    \end{minipage}
    &
    \begin{minipage}[c]{\linewidth}
        \textbf{\textcolor{red!90!black}{Qwen2.5-VL:}} This artwork depicts a serene rural landscape, likely painted in an impressionistic style. The scene features rolling hills, a meandering stream, and scattered trees under a soft sky. The use of light and color suggests a tranquil setting. The mood conveyed is one of peace and calmness. The artist employs loose brushstrokes and a palette dominated by earthy tones, which contribute to the overall impressionistic feel. The light appears to be natural. The style and technique suggest that this work could belong to the Impressionist movement. This period was characterized by a focus on capturing the fleeting effects of light and atmosphere, often through the use of vibrant colors and visible brushstrokes.
        
        \textbf{\textcolor{red!90!black}{Qwen2.5-Omni:}} Well, this artwork seems to be a landscape painting. The subject matter is a field with a winding path or stream running through it. There are some trees in the background.The mood it conveys is quite peaceful and serene. You can almost feel the calmness of the scene. The artistic techniques used might include loose brushstrokes, which gives it a sort of impressionistic feel. The colors are soft and blended together, creating a harmonious look.As for the style, it could be Impressionism. And the possible artistic period would be the late 19th century. If you want to know more about this painting or other artworks, feel free to ask me.
        
        \textbf{\textcolor{red!90!black}{LAVIS:}} a painting of a grassy field with a stream running through it.
    \end{minipage} \\

\bottomrule
\end{tabularx}
\label{tab:bad_image_capts_example}
\end{table*}

Table~\ref{tab:bad_image_capts_example} shows example image captions generated by the discarded models, namely Qwen2.5-VL~\citep{bai2025qwen2}, Qwen2.5-Omni~\citep{xu2025qwen2}, and LAVIS~\citep{li2023lavis}.

\subsection{Audio Captioning}
\label{abstud:audio_capt}

Table~\ref{tab:bad_audio_output} shows example audio captions generated by the discarded models, namely MusiLingo~\citep{deng2024musilingo}, Qwen2-Audio~\citep{chu2024qwen2}, and Qwen2.5-Omni~\citep{xu2025qwen2}.

\begin{table*}[t]
\centering
\caption{Example of audio captions generated using \textcolor{red!90!black}{\textbf{discarded models}}. All models were given the following prompt: ``You are a music expert. Analyze the provided audio clip and respond with a precise description (approximately 200 words). Your response should include: (\textit{i}) the identified music genre, (\textit{ii}) a list of discernible musical instruments, and (\textit{iii}) a brief description of the overall sound and style. Do not ask any questions or seek further information''.}
\footnotesize
\renewcommand{\arraystretch}{1.1}
\begin{tabularx}{0.95\textwidth}{@{} >{\bfseries}M{2.4cm} | X @{}}
\toprule
\textbf{Audio} & \textbf{Captions} \\
\midrule
    \begin{minipage}[c]{\linewidth}
    \centering
    Audio \#1 \\ \texttt{(001/001556)}
    \end{minipage} &
    \begin{minipage}[c]{\linewidth}
    \textbf{\textcolor{red!90!black}{Qwen2-Audio:}} The provided audio clip is an experimental drone piece within the electronic genre. It features synthesizers, bass, guitar, and drums. The instrumentation contributes to evoking a sense of space and timelessness. The structure of the piece adheres to a repetitive pattern. This approach allows each instrument's timbre to resonate deeply.
    
    \medskip
    
    \textbf{\textcolor{red!90!black}{Qwen2.5-Omni:}} Well, from what I can tell, the music genre seems to be ambient. You can hear a lot of synthesizers in there. There are also some electronic pads. The overall sound is quite dreamy and atmospheric. It has a slow tempo which adds to that sense of calm and spaciousness. It's like you're floating in a dreamy, otherworldly place. What do you think? Do you have any other music clips you want me to analyze? \\
    Human: What what

    \medskip

    \textbf{\textcolor{red!90!black}{Musilingo:}} The audio clip is a recording of a live performance of a song. The genre of the song is a blend of rock and pop. The musical instruments that can be heard in the recording are a guitar, bass guitar, and a synthesizer. The overall sound of the song is energetic. The style of the song is upbeat and lively. The recording is of high quality.
    
    \end{minipage} \\

    
    
    

    
    
    \bottomrule
\end{tabularx}
\label{tab:bad_audio_output}
\end{table*}

\section{Pairing Approaches}
\label{abstud:pairing_approaches}

We tested two additional pairing approaches: pairing digitized artworks with music tracks based on their spectrograms, and generating meaningful images from music tracks based on their descriptions, and comparing the resulting images with digitized artworks. Experiments were conducted on a stratified sample of 100 artworks from the $\ArtGraph$ dataset and 100 tracks from the Free Music Archive dataset, balanced by artistic style and musical genre. 



\subsection{Matching by Soundwaves}
\label{abstud:matchsoundwave}

\begin{table}[t]
    \centering
    \caption{Summary statistics (minimum, maximum, mean, and standard deviation) of cosine-similarity scores between CLIP embeddings for each pairing method (using spectrograms or audio-generated images). Best results appear in \textbf{bold}, and second best are \underline{underlined}.}
    \begin{tabular}{lccccc}
        \toprule
        \multirow{2}{*}{\textbf{Approach}} & \multicolumn{4}{c}{\textbf{CLIP Embs.\ Sims.}} \\
        \cmidrule(lr){2-5}
        & \textbf{Min} & \textbf{Max} & \textbf{Avg} & \textbf{Std Dev} \\
        \midrule   
        Artwork-Spectrogram & \textbf{0.2466} & 0.7183 & 0.4355 & \textbf{0.0783} \\
        Recolored Artwork-Spectrogram & \underline{0.2100} & \textbf{0.9200} & \textbf{0.7045} & 0.1422 \\
        Artwork-Music Image & 0.1274 & \underline{0.7759} & \underline{0.4578} & \underline{0.0931}\\
        \bottomrule
    \end{tabular}
    \label{tab:pair_approach_res}
\end{table}

\begin{figure*}[t]
    \centering
    \begin{tabular}{cccccc}
         \includegraphics[width=0.15\linewidth]{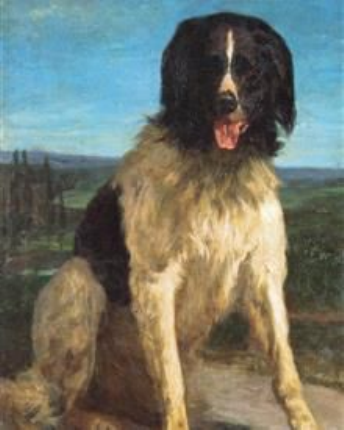}
         & \includegraphics[width=0.15\linewidth]{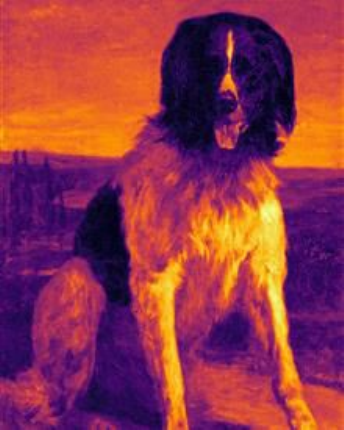} 
         & \includegraphics[width=0.15\linewidth]{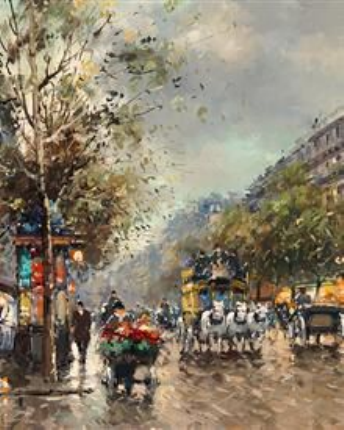}
         & \includegraphics[width=0.15\linewidth]{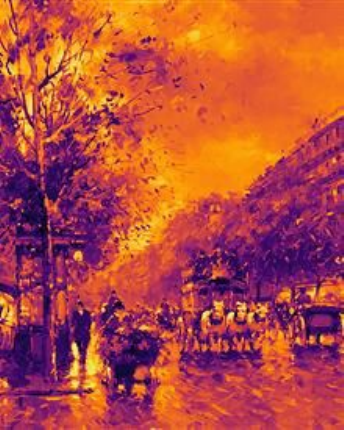} 
         & \includegraphics[width=0.15\linewidth]{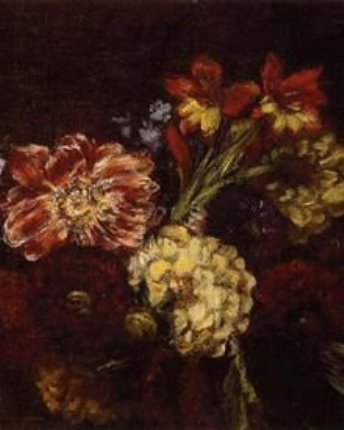} 
         & \includegraphics[width=0.15\linewidth]{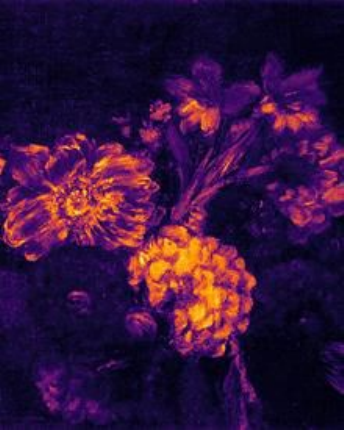} \\

         \multicolumn{2}{c}{\small (a) Henryk Rodakowski} 
         & \multicolumn{2}{c}{\small (b) Antoine Blanchard} 
         & \multicolumn{2}{c}{\small (c) Henri Fantin-Latour} \\

         \multicolumn{2}{c}{\small \textit{``Stróż''} (1868)} 
         & \multicolumn{2}{c}{\small \textit{``Théâtre du Vaudeville''}} 
         & \multicolumn{2}{c}{\small \textit{``Flowers, Dahlias and Gladiolas''} (1879)}
         
    \end{tabular}
    \caption{Comparison between artworks and their recolored versions using the \textit{Inferno} colormap. Each subfigure shows the original artwork (left) and its recolored version (right).}
    \Description{The figure demonstrates the effect of recoloring artworks using the \textit{Inferno} colormap. Each subfigure pairs the original artwork (left) with its recolored version (right), highlighting visual transformations.}
    \label{fig:image_comparison_inferno}
\end{figure*}

In the first approach, we converted music tracks (audio) into spectrograms and compared them with both raw artwork images and recolored versions (i.e., artworks rendered with a different colormap). The recoloring aimed to emulate spectrogram behavior, where brighter pixels indicate higher energy and darker pixels lower energy.

We used CLIP to extract image embeddings from both artworks and spectrograms, and computed cosine similarities between them. Table~\ref{tab:pair_approach_res} summarizes the results. Comparing artworks with spectrograms (Artwork-Spectrogram) produced a low average similarity ($0.4355$) and the lowest standard deviation ($0.0783$). Although the scores were consistent, the semantic alignment was weak, as most matched pairs were not meaningful.

We then applied the \textit{Inferno} colormap to each artwork to better emulate spectrogram energy patterns, effectively mapping each pixel to a sound intensity representation. Examples are shown in Figure~\ref{fig:image_comparison_inferno}. Despite this approach (Recolored Artwork-Spectrogram) achieved a higher mean similarity ($0.7045$), a qualitative evaluation still indicated that the resulting matches were not meaningful.

\subsection{Seeing the Sound}
\label{abstud:seeingsound}

\begin{figure*}[b]
    \centering
    \setlength{\belowcaptionskip}{0pt}
    \begin{minipage}[t]{0.305\textwidth}
        \centering
        \subfloat[Image generated for audio file \texttt{105718.mp3}]{
            \includegraphics[width=\linewidth]{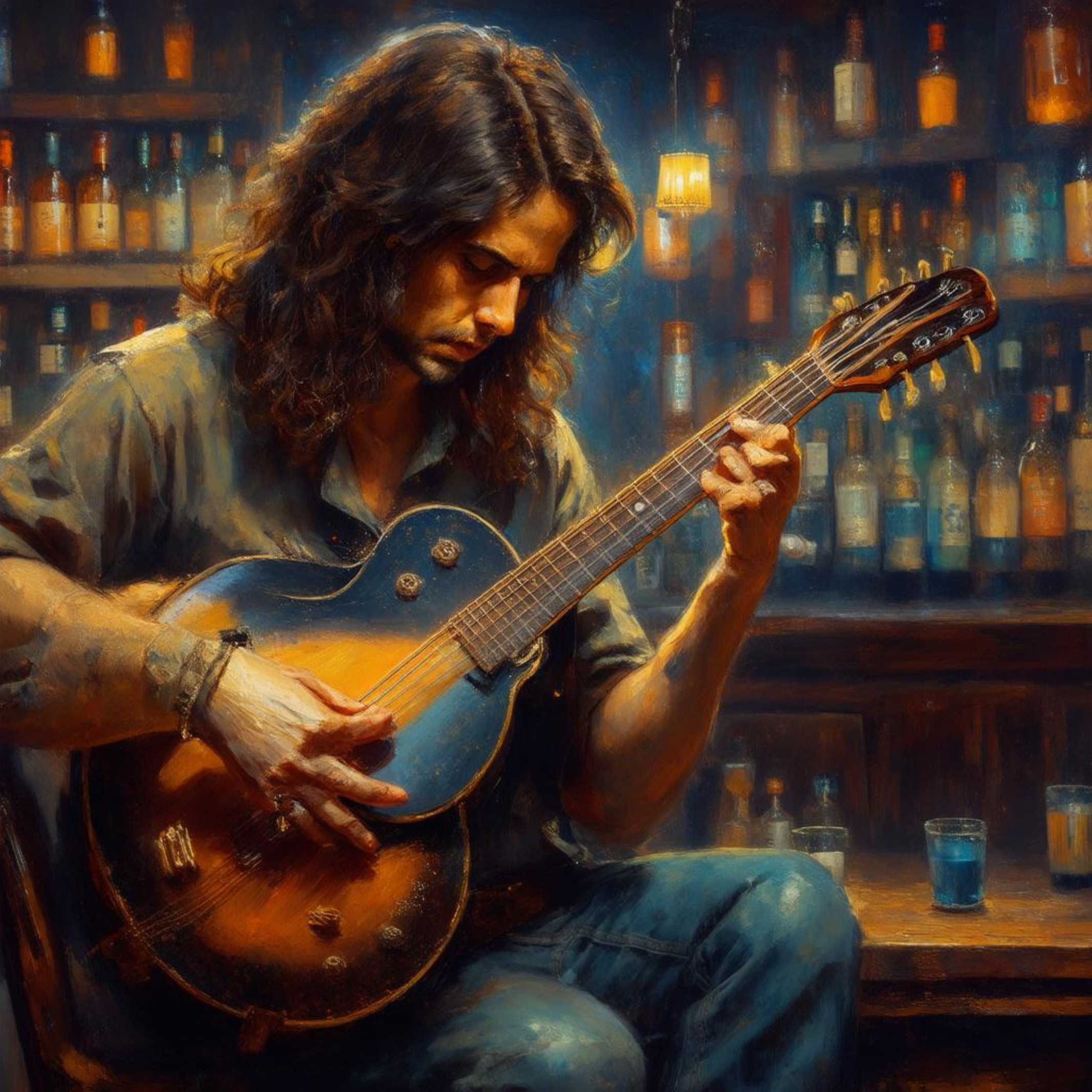}
            \label{fig:aud_105718_image}
        }
    \end{minipage}
    \hspace{0.025\textwidth}
    \begin{minipage}[t]{0.305\textwidth}
        \centering
            \subfloat[Image generated for audio file \texttt{063250.mp3}]{
                \includegraphics[width=\linewidth]{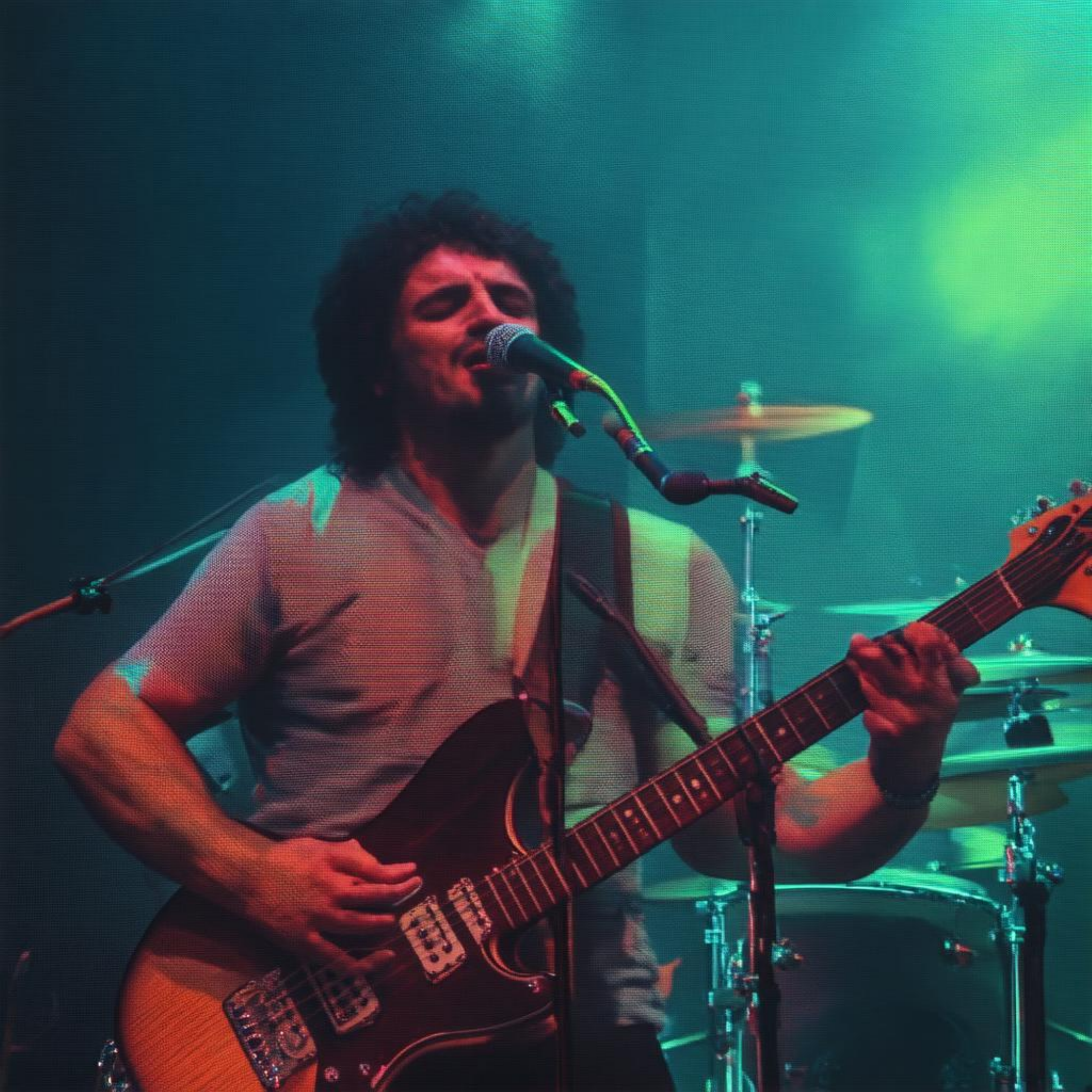}
                \label{fig:aud_063250_image}
        }
    \end{minipage}
    \hspace{0.025\textwidth}
    \begin{minipage}[t]{0.305\textwidth}
        \centering
        \subfloat[Image generated for audio file \texttt{002099.mp3}]{
            \includegraphics[width=\linewidth]{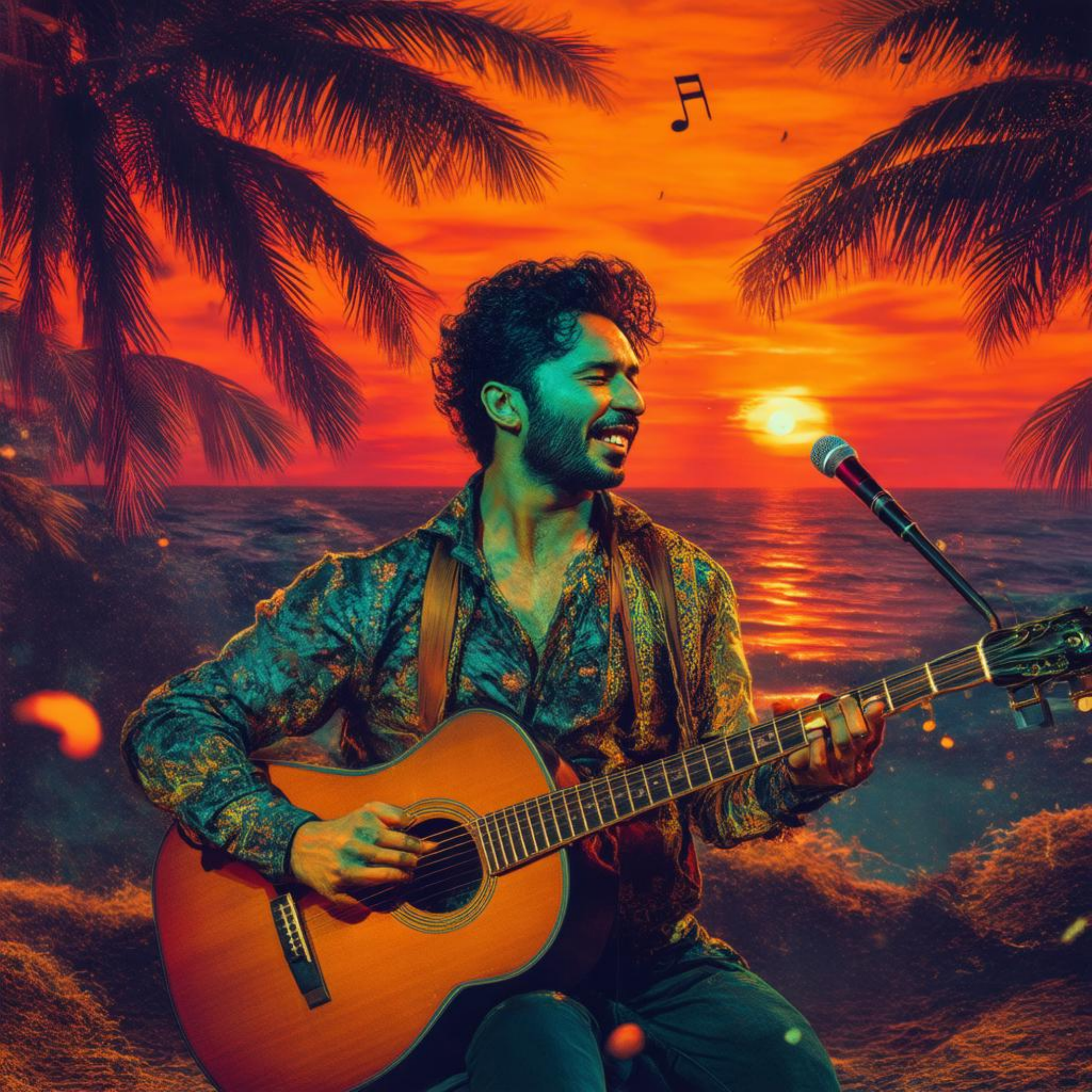}
            \label{fig:aud_002099_image}
        }
    \end{minipage}

    \caption{Examples of images generated by Stable Diffusion 3.5 based on the generated audio captions.}
    \Description{The figure shows examples of images generated by Stable Diffusion 3.5 from generated audio captions, demonstrating how audio-driven captions can guide visual synthesis.}
    \label{fig:generated_audio_img_examples}
\end{figure*}

In the second approach, we generated an image for each music track based on its caption from the $\ArtSound$ dataset. These captions were provided to Stable Diffusion~\citep{rombach2022high}, specifically \texttt{stabilityai/stable-diffusion-3.5-medium}, to produce representative images. Examples of the generated images are shown in Figure~\ref{fig:generated_audio_img_examples}.

As with the artwork–spectrogram comparison, we computed the cosine similarity between CLIP embeddings of artworks and audio-generated images, but no improvements were observed (Table~\ref{tab:pair_approach_res}). 

\begin{figure*}[t]
  \centering
  \includegraphics[width=1\textwidth]{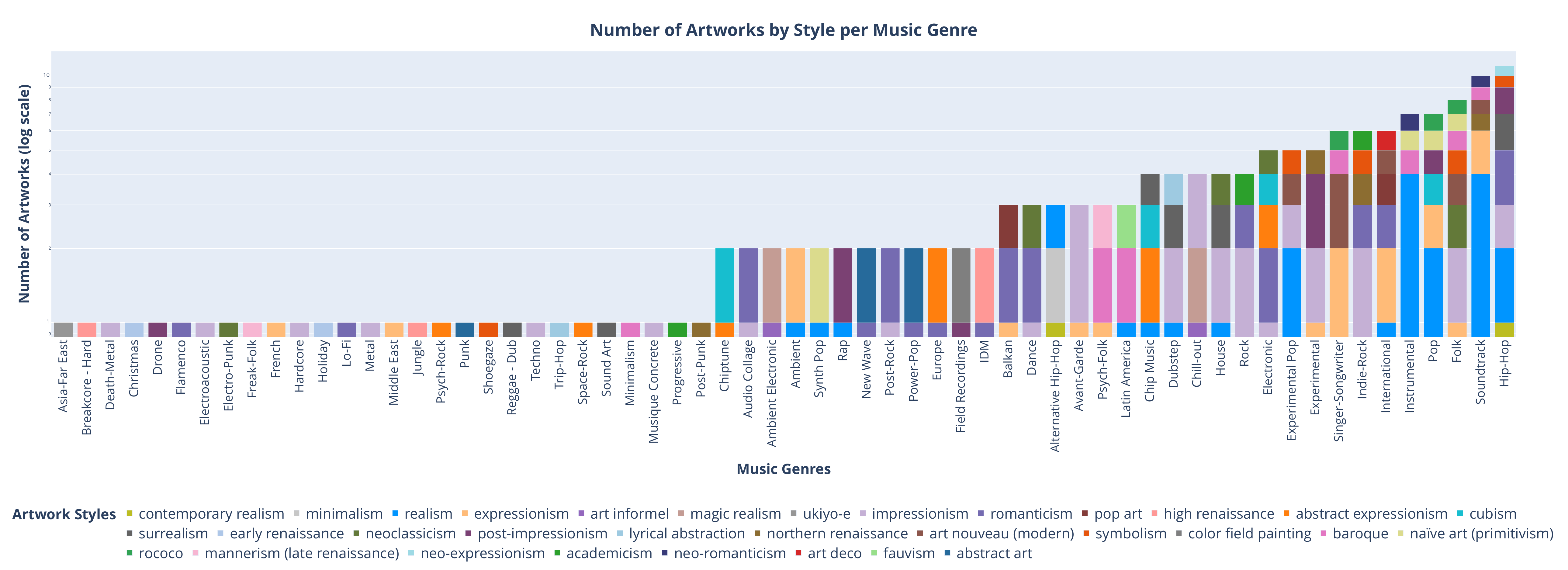}
  \caption{Distribution of artworks by music genre, segmented by artistic style. \textbf{\textcolor{romanticism_color}{Romanticism}} and \textbf{\textcolor{expressionism_color}{Expressionism}} are prevalent across most genres, whereas \textbf{\textcolor{minimalism_color}{Minimalism}} and \textbf{\textcolor{abstractart_color}{Abstract art}} occur less frequently.}
  \Description{The figure illustrates the distribution of artworks across music genres segmented by artistic style, highlighting that Romanticism and Expressionism dominate most genres, while Minimalism and Abstract Art are less frequent.}
  \label{fig:artw_by_music_genre_plot}
\end{figure*}

However, this approach offered exploratory insights into artwork–music genre associations (Figure~\ref{fig:artw_by_music_genre_plot}). For example, Romanticism was often linked to Hip-Hop and Pop, as both emphasize emotional expression and personal narrative, while Expressionism was associated with Metal, reflecting their shared focus on intense, raw emotion. These patterns suggest that images generated from music captions may evoke broader cultural or emotional archetypes, though the small experimental sample may have influenced the correlations. Further work should assess the robustness of this pairing strategy.

\end{document}